%% file: main.tex
\documentclass[11pt]{article}

\usepackage[T1]{fontenc}
\usepackage{lmodern}
\usepackage[margin=1in]{geometry}
\usepackage{microtype}
\usepackage{amsmath,amssymb,bm}
\usepackage{booktabs}
\usepackage{graphicx}
\usepackage{subcaption}
\usepackage{array}
\usepackage{tabularx}
\usepackage{multirow}
\usepackage{xcolor}
\usepackage{natbib}
\usepackage{url}
\usepackage[hidelinks]{hyperref}

\input{math_commands.tex}

\hypersetup{
  pdftitle={From Memory Access to Evidence Sufficiency: Diagnosing and Addressing Demand--Evidence Misalignment},
  pdfauthor={Shiyao Liu, Yixing Fan, Ruiqing Zhang, Yan Xiaohui, Jiafeng Guo, Xueqi Cheng}
}

\title{From Memory Access to Evidence Sufficiency: Diagnosing and Addressing Demand--Evidence Misalignment}

\author{}
\date{}

\begin{document}

\maketitle

\vspace{-2.8em}

\begin{center}

\makebox[\textwidth][c]{%
\begin{tabular*}{1.08\textwidth}{@{\extracolsep{\fill}}ccc}

\begin{minipage}[t]{0.338\textwidth}
\centering
{\Large Shiyao Liu}\\[2pt]
{\small
Institute of Computing Technology,\\
Chinese Academy of Sciences\\
Beijing, China\\
liushiyao23s@ict.ac.cn}
\end{minipage}
&
\begin{minipage}[t]{0.338\textwidth}
\centering
{\Large Yixing Fan$^{\dagger}$}\\[2pt]
{\small
Institute of Computing Technology,\\
Chinese Academy of Sciences\\
Beijing, China\\
fanyixing@ict.ac.cn}
\end{minipage}
&
\begin{minipage}[t]{0.338\textwidth}
\centering
{\Large Ruiqing Zhang}\\[2pt]
{\small
Institute of Computing Technology,\\
Chinese Academy of Sciences\\
Beijing, China\\
zhangruqing@ict.ac.cn}
\end{minipage}

\end{tabular*}%
}

\vspace{1.35em}

\makebox[\textwidth][c]{%
\begin{tabular*}{1.08\textwidth}{@{\extracolsep{\fill}}ccc}

\begin{minipage}[t]{0.338\textwidth}
\centering
{\Large Yan Xiaohui}\\[2pt]
{\small
Huawei Technologies Co., Ltd.\\
China\\
yanxiaohui2@huawei.com}
\end{minipage}
&
\begin{minipage}[t]{0.338\textwidth}
\centering
{\Large Jiafeng Guo}\\[2pt]
{\small
Institute of Computing Technology,\\
Chinese Academy of Sciences\\
Beijing, China\\
guojiafeng@ict.ac.cn}
\end{minipage}
&
\begin{minipage}[t]{0.338\textwidth}
\centering
{\Large Xueqi Cheng}\\[2pt]
{\small
Institute of Computing Technology,\\
Chinese Academy of Sciences\\
Beijing, China\\
cxq@ict.ac.cn}
\end{minipage}

\end{tabular*}%
}

\vspace{0.8em}

{\small $^{\dagger}$ Corresponding authors.}

\end{center}

\vspace{0.5em}


\input{sec/0_abstract}

\input{sec/1_introduction}
\input{sec/5_related_works}
\input{sec/2_MemPath}
\input{sec/3_BiDE-Mem}
\input{sec/4_experiments}
\input{sec/6_discussion_limitations}
\input{sec/6_conclusion}

\input{sec/7_Ap}


\bibliographystyle{plainnat}
\bibliography{references}

\end{document}

%% file: math_commands.tex
\usepackage{amsmath,amsfonts,bm}
\usepackage{multirow}

\def\eqref#1{equation~\ref{#1}}

\def\1{\bm{1}}

\DeclareMathAlphabet{\mathsfit}{\encodingdefault}{\sfdefault}{m}{sl}
\SetMathAlphabet{\mathsfit}{bold}{\encodingdefault}{\sfdefault}{bx}{n}



%% file: sec/0_abstract.tex
\begin{abstract}
Long-term memory systems are typically optimized to preserve and retrieve
records relevant to a later query. Yet relevance is an item-level property,
whereas answering is a set-level requirement: the retrieved evidence must
jointly establish the facts, relations, temporal conditions, and state
constraints required by the question. We characterize this challenge as
demand--evidence misalignment and introduce MemPath, an interventional
diagnostic framework that traces the transformation from historical information
to answer-supporting evidence across the memory lifecycle. Across
representative systems, MemPath reveals that improving individual memory stages
does not necessarily ensure sufficient evidence for answering, as failures can
arise from insufficient access, incomplete demand coverage, or ineffective
evidence utilization. Guided by this diagnosis, we propose BiDE-Mem, a
demand-guided evidence construction framework that represents questions as
typed demands and uses these requirements to guide retrieval and evidence
organization. We formalize the demand--evidence support relation; in the
evaluated compact path, demand awareness is realized through provenance-aware
evidence organization and Reader guidance rather than a separate generative
four-state verification call. BiDE-Mem improves accuracy by 6.23 points on
LoCoMo and achieves 91.80\% overall accuracy on LongMemEval, with strong results on LoCoMo multi-hop and temporal questions. Supplementary Qwen-Reader
evaluations report the highest overall accuracy among seven evaluated system
configurations on both LoCoMo and LongMemEval-S under two judging protocols.
Ablations and evidence-level analyses indicate that query representation,
candidate retrieval, and evidence presentation all affect answer performance,
beyond memory access alone. Our results highlight that effective long-term
memory requires not only accessing relevant information, but also constructing
evidence that addresses the requirements of the current question.
\end{abstract}

%% file: sec/1_introduction.tex
\section{Introduction}
\label{sec:introduction}

Long-term memory systems are evaluated by whether they can retrieve information from past interactions and use it to answer a question. This view, however, obscures a critical distinction. Retrieval relevance is an item-level property: a memory may mention the same person, event, or topic as the question. Answering is a set-level requirement: the available evidence must jointly establish the facts, relations, temporal conditions, and state constraints required by the question. Consequently, retrieving relevant memories does not imply that the resulting context is sufficient for answering.

\begin{figure}[ht]
    \centering
    \includegraphics[width=\linewidth]{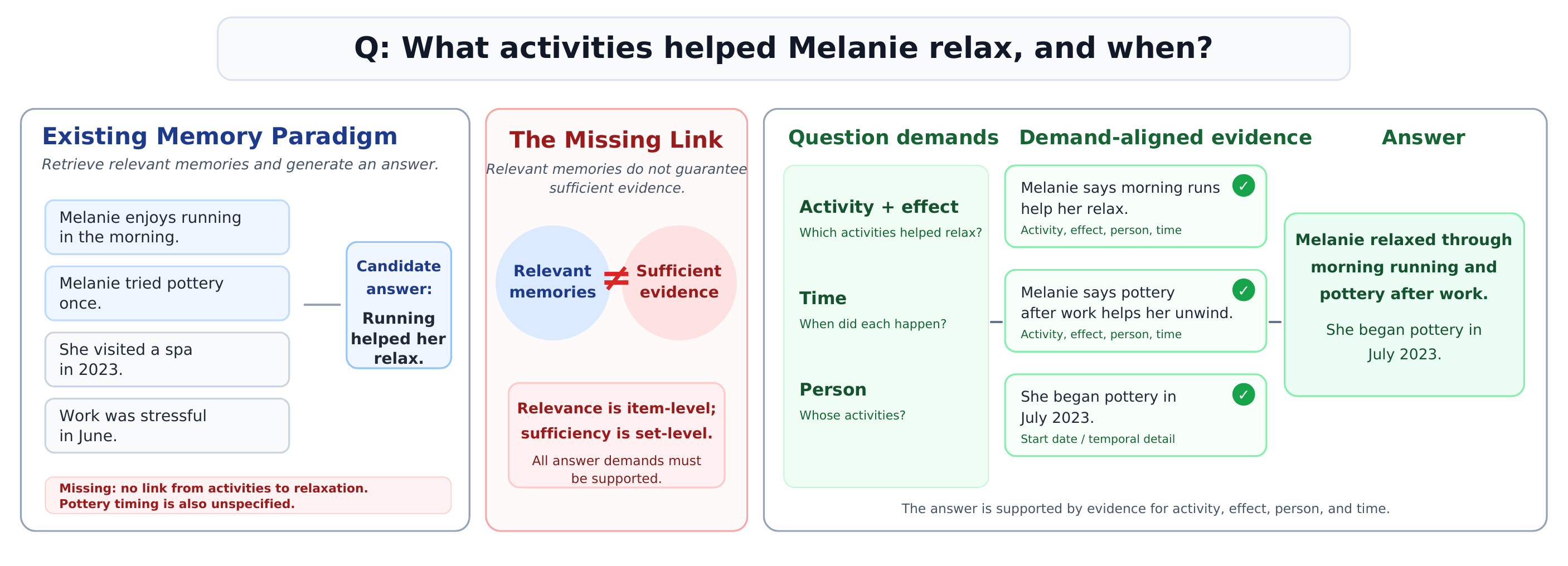}
    \caption{Illustrative example: relevant memories can omit the relations and temporal details required to answer a question.}
    \label{fig:motivation}
\end{figure}

Consider the question in Figure~\ref{fig:motivation}: ``What activities helped Melanie relax and when?'' Memories mentioning Melanie's running, pottery, or stressful work may all be topically relevant. Yet a complete answer requires more than retrieving activity-related records. The evidence must identify the activities that produced the requested effect, connect each activity to Melanie, and establish the corresponding temporal information. We refer to these answer conditions as \emph{question demands}, denoted by $D(Q)$. Given a retrieved evidence set $E$, high relevance between individual records and $Q$ does not guarantee that $E \models D(Q)$. We call this discrepancy \emph{demand--evidence misalignment}.

Demand--evidence misalignment cannot be fully studied using only end-to-end accuracy. Memory systems can fail for multiple reasons: information may be lost during memory construction, stored evidence may become inaccessible, retrieval may omit required support, budgeted selection may discard critical dependencies, or the model may fail to effectively use otherwise sufficient evidence.

Existing long-term memory benchmarks such as LoCoMo~\citep{maharana2024evaluating} and LongMemEval~\citep{wu2025longmemeval} evaluate memory under multi-session interaction, temporal reasoning, knowledge updates, and long-range dependencies, but primarily measure final answer quality. Although LongMemEval examines design choices in indexing, retrieval, and reading, the root causes and propagation of individual failures across the memory pathway remain underexplored.

Recent methods have begun to address limitations beyond similarity-based retrieval. EviMem~\citep{li2026evimem} studies evidence-gap-driven iterative retrieval; MemRetriever~\citep{jiang2026memretriever} learns multi-step search and reflection; DeferMem~\citep{yin2026defermem} studies query-time evidence distillation; and GraphMemix~\citep{li2026graphmemix} constructs query-aware evidence forests under a context budget. These methods establish that evidence acquisition and organization are active research problems. Our focus is complementary: MemPath provides a lifecycle-level diagnostic view with source accessibility, demand coverage, and oracle evidence intervention, while BiDE-Mem uses typed question demands as a shared reference for retrieval and Reader-facing evidence organization.

We therefore introduce \textbf{MemPath}, an interventional diagnostic framework that traces how historical information is transformed into answer-supporting evidence. Using question demands as a shared reference, MemPath evaluates source accessibility, demand coverage, and answer recovery under oracle evidence interventions. Applying MemPath to memory systems reveals that accessibility, sufficiency, and utilization are distinct bottlenecks: one system has retrieval-stage coverage exceeding $93\%$ of annotated sources but achieves only $85.3\%$ answer accuracy, while combined oracle evidence recovers $54.8\%$ of valid initially incorrect cases, with different effects across systems. These findings show that retrieving relevant memories or adding more context does not guarantee success; the available information must be organized into evidence that satisfies the question's demands.

Motivated by this diagnosis, we propose \textbf{BiDE-Mem}, a long-term memory framework with bidirectional demand--evidence alignment. BiDE-Mem represents questions as factual, relational, temporal, state, and constraint demands, operationalizes these requirements through QueryFrame for demand-conditioned retrieval, and organizes retrieved records into provenance-preserving evidence blocks and Reader guidance. We formalize the relationship between each demand and its supporting evidence as a support mapping. In the evaluated compact path, this relationship is conveyed through the organized Reader-facing context; a separate generative four-state verifier is not part of the reported Full configuration. BiDE-Mem improves overall accuracy by $6.23$ points on LoCoMo and achieves $91.80\%$ overall accuracy on LongMemEval, with strong results on LoCoMo multi-hop and temporal questions. Supplementary Qwen-Reader evaluations show the highest overall accuracy among seven evaluated system configurations on LoCoMo and LongMemEval-S under two judging protocols, with system-specific memory inputs and retrieval procedures retained. Ablations and evidence-level analyses on LoCoMo highlight the contributions of explicit demand modeling, retrieval, and evidence organization beyond memory access alone.

Our contributions are:
\begin{itemize}
    \item We identify and formulate demand--evidence misalignment in long-term agent memory, separating memory relevance from evidence sufficiency.
    \item We introduce MemPath, an interventional diagnostic framework that traces the transformation from historical interactions to answer-supporting evidence and localizes failures through source accessibility, demand coverage, and oracle evidence interventions.
    \item We propose BiDE-Mem, a memory framework based on bidirectional demand--evidence alignment, which uses question demands to guide memory access and evidence construction while formalizing demand--evidence support relationships and organizing the Reader-facing evidence accordingly.
    \item We evaluate BiDE-Mem on LoCoMo and LongMemEval, demonstrating strong performance across both benchmarks, with supplementary shared-Reader evaluations, reported component comparisons, and evidence-level analyses on LoCoMo.
\end{itemize}

%% file: sec/5_related_works.tex
\section{Related Work}
\label{sec:related_work}

\paragraph{Long-term memory for LLM agents.}
Long-term memory systems extend language-model agents beyond a fixed context window by retaining and reusing information from prior interactions. Early work studies memory as an interaction history that can be retrieved or summarized, while subsequent systems introduce persistent stores, reflection, hierarchical organization, and structured representations \citep{park2023generative,packer2023memgpt,zhong2024memorybank,chhikara2025mem0,tan2025prospect,xu2025amem,wang2025mirix,hu2026evermemos,tang2026mnemis}. More recent architectures explore multi-graph organization, structured long-horizon memory, retrieval-driven reconsolidation, and memory-efficient processing \citep{jiang2026magma,xu2026structmem,song2026retrieval,liu2026simplemem,fang2026lightmem}. These studies address how memories are formed, represented, updated, and accessed. Our focus is complementary: given a question, do the accessible records jointly support all of its answer requirements?

\paragraph{Retrieval and evidence construction.}
Retrieval-augmented generation and dense passage retrieval provide mechanisms for accessing relevant information from large collections \citep{lewis2020retrieval,karpukhin2020dense}. Multi-hop question answering and complementary-evidence research show that answering may require combining multiple records and preserving their relationships \citep{yang2018hotpotqa,welbl2018constructing,mou2021complementary}. Recent memory methods move beyond one-shot similarity retrieval: EviMem studies evidence-gap-driven iterative retrieval, MemRetriever learns multi-step search and reflection, and DeferMem studies query-time evidence distillation \citep{li2026evimem,jiang2026memretriever,yin2026defermem}. GraphMemix constructs query-aware evidence forests under a context budget, while RaMem emphasizes contextual reinstatement for agentic memory \citep{li2026graphmemix,yang2026ramem}. These methods make evidence quality and organization explicit. BiDE-Mem differs in using a typed representation of question demands as a shared reference for retrieval and Reader-facing evidence organization; MemPath separately evaluates accessibility, annotated demand coverage, and evidence intervention across heterogeneous memory systems.

\paragraph{Evaluation and diagnosis of long-term memory.}
LoCoMo evaluates conversational memory across multiple reasoning categories, and LongMemEval tests long-term information access, temporal reasoning, updates, and multi-session dependencies \citep{maharana2024evaluating,wu2025longmemeval}. A-TMA analyzes state-aware memory failures across the memory process, complementing benchmark-level answer evaluation \citep{shi2026tma}. These benchmarks and analyses motivate more detailed evaluation, but end-to-end answer accuracy alone cannot identify whether a failure is due to inaccessible sources, incomplete support for a question demand, or ineffective use of available evidence. MemPath addresses this diagnostic gap through a common lifecycle view and oracle evidence interventions.

%% file: sec/2_MemPath.tex
\section{MemPath: Diagnosing Demand--Evidence Misalignment}
\label{sec:mempath}

Existing long-term memory systems optimize different stages of memory utilization, including memory construction, organization, retrieval, and context composition \citep{park2023generative,packer2023memgpt,zhong2024memorybank,chhikara2025mem0,yang2026ramem,wang2025mirix}. However, final answer accuracy does not reveal whether failures originate from missing information, incomplete evidence construction, or ineffective evidence utilization. We introduce \textbf{MemPath}, an interventional diagnostic framework that traces how historical information is transformed into answer-supporting evidence and identifies where demand--evidence misalignment emerges.

\subsection{Canonical Memory Pathway}

MemPath represents long-term memory utilization as a sequence of transformations:

\begin{equation}
H \rightarrow M \rightarrow O \rightarrow C_Q \rightarrow E_Q \rightarrow A_Q ,
\end{equation}

where $H$ denotes historical interactions, $M$ denotes constructed memory, $O$ denotes memory organization, $C_Q$ denotes the retrieved candidate pool, $E_Q$ denotes the assembled evidence context, and $A_Q$ denotes the final answer. This pathway serves as a canonical interface for heterogeneous memory systems rather than assuming identical internal structures. For each system, observable artifacts are mapped to corresponding stages, while unavailable stages are treated as unobserved.

To evaluate whether information available at each stage can support answering, MemPath introduces Question Demands as a shared reference. For a question $Q$, we define:

\begin{equation}
D(Q)=\{d_1,\ldots,d_m\},
\end{equation}

where each demand $d_i$ represents an answer requirement involving required facts, relations, temporal conditions, states, or constraints. Unlike retrieval relevance, which evaluates individual records, Question Demands characterize whether the collected evidence jointly satisfies the conditions required for answering. Conceptually, a demand is satisfied when the available evidence supports the corresponding answer condition.

Based on this representation, MemPath separates three aspects that are conflated by end-to-end accuracy: whether required information is accessible, whether assembled evidence satisfies question demands, and whether improving evidence can recover incorrect answers.

\subsection{Demand--Evidence Diagnosis}

MemPath analyzes demand--evidence misalignment from three complementary perspectives.

\paragraph{Lifecycle Accessibility.}

We first examine whether required information successfully propagates through the memory pathway. We measure accessibility at different stages by tracking whether supporting information reaches the retrieved candidates and final evidence context. Table~\ref{tab:lifecycle} summarizes the memory organization, retrieval behavior, evidence construction, and final performance of representative systems. Detailed metric definitions, source annotation procedures, and implementation details are provided in Appendix~\ref{app:mempath}.

\begin{table*}[ht]
\centering
\small
\caption{
Lifecycle-level analysis of long-term memory systems.
}
\label{tab:lifecycle}

\resizebox{\textwidth}{!}{
\begin{tabular}{lcccccc}
\toprule

System &
Memory Organization &
Retrieval Budget &
Retrieval Coverage &
Evidence Budget &
Prompt Coverage &
Accuracy
\\

\midrule

Mnemis &
Graph/event memory &
62.33 units &
93.68\% &
62.33 units &
95.85\% &
85.26
\\

MIRIX &
Episodic memory + tools &
39.95 items &
80.80\% &
Tool evidence &
78.77\% &
83.51
\\

Mem0 &
Semantic memory &
60.00 items &
90.37\% &
60.00 items &
90.17\% &
80.39
\\

EverOS &
Episode memory &
10 episodes &
75.96\% &
Episode context &
75.96\% &
80.19
\\

ReadAgent &
Gist + page hierarchy &
2.44 pages &
73.63\% &
48.96 units &
99.87\% &
74.48
\\

MemoryBank &
Retention entries &
Top-5 entries &
25.85\% &
5 entries &
25.93\% &
29.61
\\

\bottomrule
\end{tabular}
}
\end{table*}

The results reveal diverse accessibility patterns across systems. MemoryBank suffers primarily from insufficient information access, with retrieval and prompt coverage of only $25.85\%$ and $25.93\%$, respectively, which limits the amount of usable evidence available for answering. In contrast, ReadAgent substantially increases information availability through hierarchical expansion and achieves $99.87\%$ prompt coverage, but its answer accuracy remains at $74.48\%$, showing that evidence availability alone does not guarantee successful answering. Meanwhile, Mnemis retrieves over $93\%$ of annotated supporting information and achieves high prompt coverage ($95.85\%$), yet still obtains only $85.26\%$ accuracy, indicating that accessible information may still require effective evidence organization and utilization before answer generation.

\paragraph{Demand Completeness.}

Lifecycle accessibility indicates whether annotated supporting information reaches later stages of the memory pathway. We measure \emph{source-grounded demand coverage} using the annotated support set $A_i$ for each demand $d_i$. For each demand, let $S_i(E_Q)$ be the set of its annotated supporting sources \emph{retained in the observed evidence context}. For uncompressed source-preserving records, source-ID presence identifies the retained original content. For compressed or structured memories, we use semantic mapping to establish whether the retained content preserves the information required by $d_i$. The demand index matters because the same source may preserve support for one requirement but not another. The all-support criterion remains

\begin{equation}
\operatorname{Cov}_{\mathrm{src}}(E_Q,D(Q))
=\frac{1}{|D(Q)|}\sum_{d_i\in D(Q)}
\mathbb{I}[A_i\subseteq S_i(E_Q)].
\label{eq:demand_coverage}
\end{equation}

A question is termed \emph{demand-complete} in the diagnostic plots when all its demands satisfy the annotated all-support criterion. The metric evaluates coverage of the required supporting information, with representation-aware matching for compressed evidence; it does not require the Reader to produce a correct answer. Appendix~\ref{app:diagnostic_protocol} details the coverage rules.

\begin{figure}[t]
    \centering
    \begin{subfigure}[t]{0.48\linewidth}
        \centering
        \includegraphics[width=\linewidth]{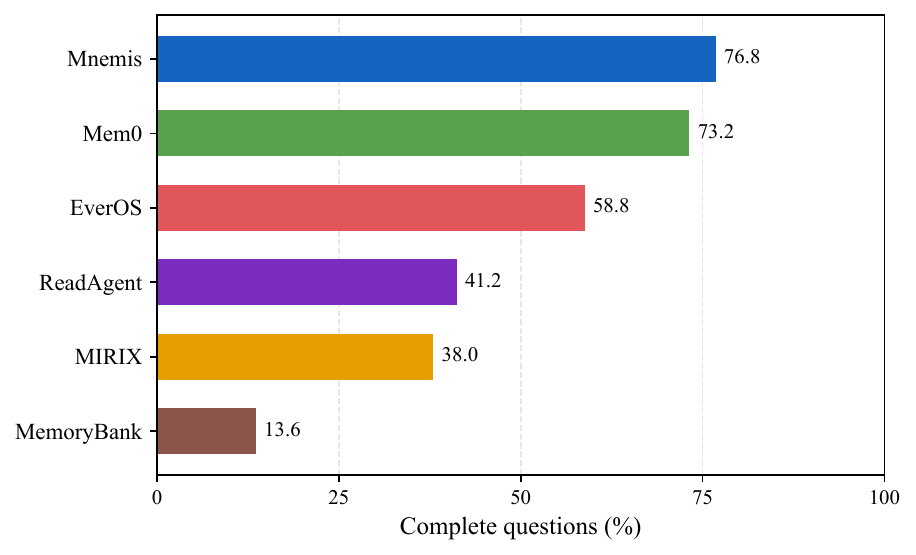}
        \caption{Question-level demand completeness.}
        \label{fig:diagnosis_demands}
    \end{subfigure}
    \hfill
    \begin{subfigure}[t]{0.48\linewidth}
        \centering
        \includegraphics[width=\linewidth]{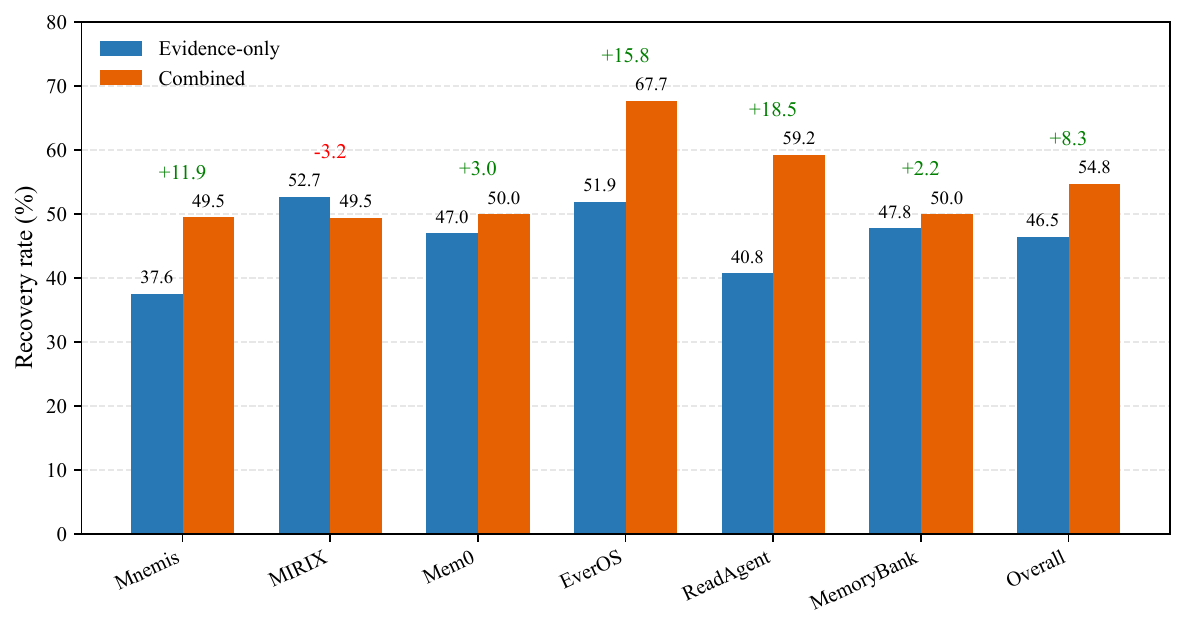}
        \caption{Oracle evidence intervention recovery.}
        \label{fig:diagnosis_interventions}
    \end{subfigure}
    \caption{
    Demand--evidence diagnosis with MemPath.
    (a) Question-level source-grounded completeness on 250 diagnostic questions containing 425 annotated demands.
    (b) Reported answer recovery from 869 valid system--question error cases per intervention setting; the aggregate ledger alone does not establish paired case identifiers.
    }
    \label{fig:diagnosis}
\end{figure}

Figure~\ref{fig:diagnosis_demands} reports the percentage of demand-complete questions on the reviewed diagnostic subset of 250 questions containing 425 Question Demands. The results distinguish aggregate source accessibility from coverage of the annotated demand support sets. For ReadAgent, 103/250 diagnostic questions (41.2\%) meet the all-support criterion. Its 99.87\% prompt-stage source coverage in Table~\ref{tab:lifecycle} uses a different denominator and therefore answers a different diagnostic question.

\paragraph{Evidence Intervention.}

MemPath evaluates whether evidence improvements can recover initially incorrect answers. We consider two oracle interventions: \emph{Evidence-only}, which replaces the original context with reviewed supporting evidence, and \emph{Combined}, which integrates oracle evidence with the original context. Both keep the reader and answer-generation procedure unchanged.

Unlike question-level source-grounded completeness, intervention analysis examines initially incorrect system--question cases. The record contains 915 errors and 869 valid cases per intervention setting. Evidence-only recovers 404 cases ($46.49\%$), while Combined recovers 476 ($54.78\%$). The aggregates do not establish identical case membership or exclusion reasons; see Appendix~\ref{app:diagnostic_protocol}.

Figure~\ref{fig:diagnosis_interventions} shows that evidence intervention recovers many incorrect answers, with varying effects across systems. While Combined achieves higher overall recovery, it performs worse than Evidence-only for MIRIX, showing that additional context is not uniformly beneficial. These findings suggest that failures depend on evidence composition, not availability alone.

\subsection{Diagnostic Findings}

MemPath reveals that improving individual memory stages does not necessarily produce sufficient evidence for answering. Retrieval accessibility, context availability, demand completeness, and final accuracy capture different aspects of memory utilization, but none alone determines whether the assembled evidence is sufficient for answering.

Across systems, failures may arise from inaccessible information, incomplete answer-bearing content, or ineffective evidence use. Source-level diagnostics cannot alone certify semantic satisfaction, while oracle interventions show that changing evidence composition can recover some incorrect answers. These observations motivate demand-aware retrieval and evidence organization, with answer requirements used to structure the Reader-facing context.

%% file: sec/3_BiDE-Mem.tex
\section{BiDE-Mem: Bidirectional Demand--Evidence Alignment}
\label{sec:bide}

\subsection{Overview}
\label{sec:bide_overview}

The diagnosis in Section~\ref{sec:mempath} shows that retrieving relevant memories does not guarantee sufficient evidence for answering. BiDE-Mem addresses this gap by using explicit answer requirements to guide evidence retrieval, organization, and utilization. Historical interactions are processed into a reusable memory representation independently of the current question; question-specific demands are introduced when the system needs to retrieve evidence and answer. Additional implementation details and prompt contracts are provided in Appendix~\ref{app:bide_details} and~\ref{app:prompt_contract}.

Given a question $Q$, BiDE-Mem constructs an explicit demand representation to guide evidence acquisition and organization, and conceptually associates the selected evidence back with these demands before answer generation:

\begin{equation}
Q\rightarrow D(Q)\rightarrow C_Q\rightarrow (E_Q,Z_Q)\rightarrow A_Q ,
\label{eq:bide_pipeline}
\end{equation}

where $D(Q)$ denotes Question Demands, $C_Q$ the retrieved candidate memories, $E_Q$ the constructed evidence package, $Z_Q$ the demand--evidence support mapping, and $A_Q$ the final answer.

\begin{figure*}[ht]
    \centering
    \includegraphics[width=\textwidth]{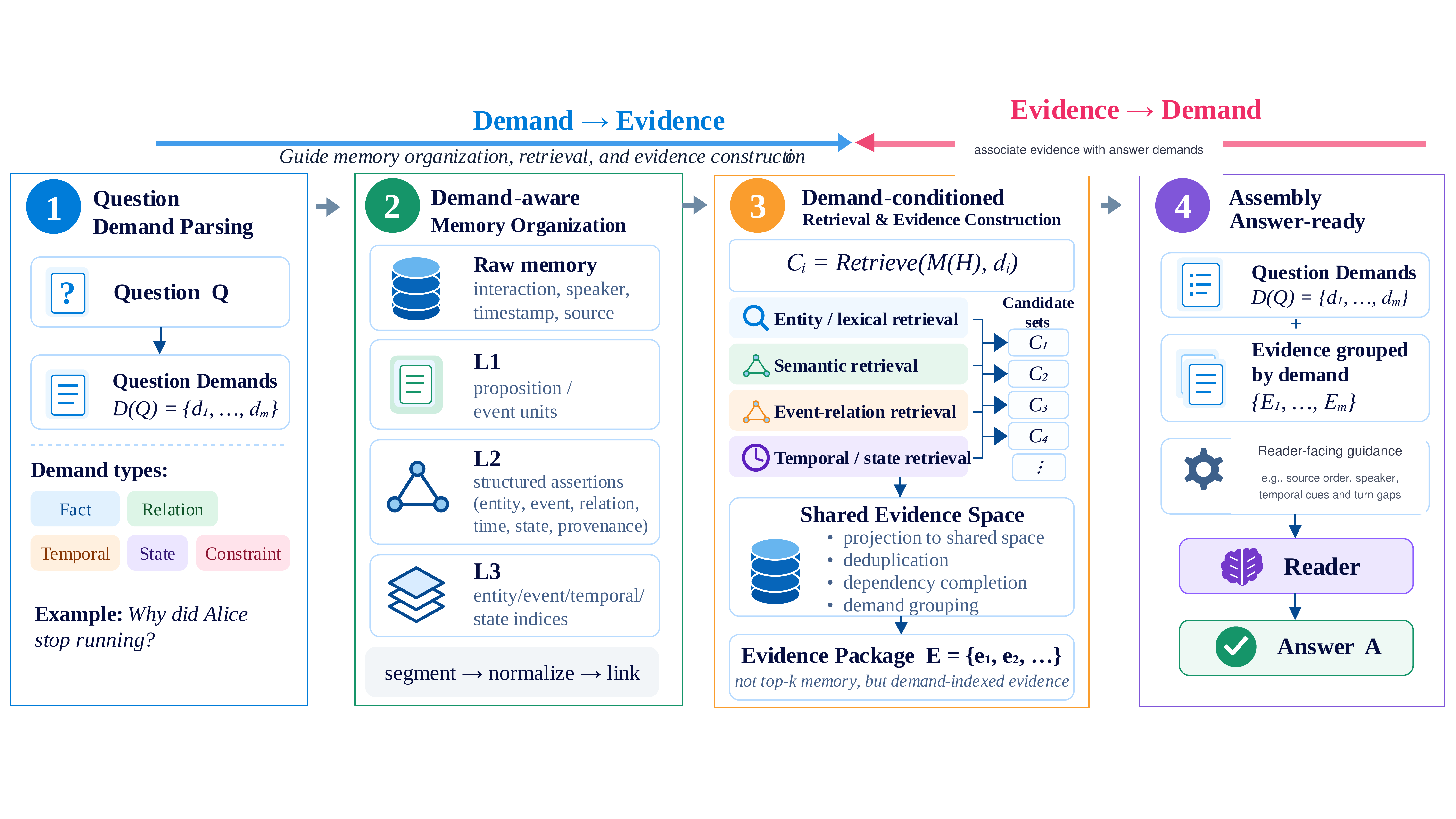}
    \caption{Overview of BiDE-Mem.}
    \label{fig:bide}
\end{figure*}

Figure~\ref{fig:bide} illustrates the two directions: demands constrain what evidence is sought and retained, while the resulting evidence is associated back with the requirements it can support. The resulting evidence presentation, rather than an unordered list of retrieved memories, is supplied to a single answer-generation pass.

\subsection{Question Demand Representation}
\label{sec:bide_demands}

Following MemPath, BiDE-Mem treats Question Demands as answer-oriented conditions rather than retrieval objectives. For a question $Q$, the required conditions are represented as

\begin{equation}
D(Q)=\{d_1,\ldots,d_m\}.
\label{eq:demand_set}
\end{equation}

Each demand $d_i$ specifies an answer requirement involving entities or events, relations, temporal conditions, states, or constraints. We categorize them into five conceptual types: \emph{Fact}, \emph{Relation}, \emph{Temporal}, \emph{State}, and \emph{Constraint}. A demand may require evidence from multiple records, and one record may support more than one demand; this makes demand satisfaction different from matching a single query to a single memory.

At inference time, BiDE-Mem operationalizes the demands through \emph{QueryFrame}, which records target entities, answer roles, relation and event cues, and temporal or state constraints for retrieval and evidence organization. QueryFrame is an operational query representation, not a one-to-one encoding of the five analytical demand types; its schema and construction procedure are provided in Appendix~\ref{app:bide_details}.

\subsection{Demand-guided Evidence Acquisition}
\label{sec:bide_retrieval}

Demand-guided retrieval aims to collect evidence for the conditions represented by $D(Q)$, instead of relying on similarity to the original question alone. Conceptually, it can be written as

\begin{equation}
C_i=\operatorname{Retrieve}(\mathcal{M},d_i),
\qquad
C_Q=\bigcup_{i=1}^{m}C_i ,
\label{eq:bide_retrieval}
\end{equation}

where $C_Q$ denotes the candidate pool for question $Q$. This equation describes demand-level coverage; it does not require a separate retrieval call or a dedicated channel for every demand.

In the implementation, QueryFrame conditions queries across six complementary retrieval channels: \emph{Clause}, \emph{Occurrence}, \emph{Structural}, \emph{Multivector}, \emph{SPLADE}, and \emph{Typed-L2}. The channels offer lexical, semantic, and structured access to stored information. In particular, entity, relation, event, and temporal fields can inform queries over structured assertions, while other channels recover relevant textual evidence. Demand types therefore influence query construction and evidence matching, rather than activating channels through a fixed type-to-channel rule.

Candidates from these channels are mapped to shared evidence records, fused, deduplicated, and reranked before selecting the final Top-$K$ under the evidence budget. Source identifiers and structural associations are retained so that evidence retrieved through different channels can be combined for multi-record reasoning. Channel configurations and selection details are given in Appendix~\ref{app:bide_details}.

\subsection{Demand-guided Evidence Organization}
\label{sec:bide_assembly}

Retrieved candidates do not constitute sufficient evidence. BiDE-Mem therefore organizes them into an answer-ready evidence package guided by Question Demands. The reusable memory representation preserves information at complementary levels: textual propositions (L1), entity/event/relation assertions with temporal and source attributes (L2), and access indices (L3). These representations are prepared from historical interactions and are not reconstructed from the test question.

After retrieval, the Top-$K$ records are assembled into compact contextual evidence blocks. The assembly retains original content and provenance where available, including speaker, timestamp, source identifier, and session/turn order. It reduces duplicate records while keeping event continuity, temporal order, and distinct state updates explicit. Reader-facing guidance then presents the question together with this organized evidence rather than a flat sequence of retrieved hits.

The demand--evidence mapping $Z_Q$ is a formal representation of how evidence relates to each answer requirement. Its semantic states are \emph{satisfied}, \emph{partial}, \emph{conflicting}, and \emph{unmet}. In the evaluated compact Reader path, demand-relevant support is conveyed through the retained records, provenance, context grouping, and Reader guidance; the path does not serialize a separate four-state $Z_Q$ object or invoke a generative verification call. The explicit ALIGN/VERIFY pilot is documented separately in Appendix~\ref{app:evidence_validation} and is not included in the reported Full results.

The normal online path uses one generative LLM call to construct QueryFrame and one to generate the final answer. Retrieval, reranking, evidence assembly, and support organization incur additional non-generative computation but do not require iterative answer refinement. Finally, the Reader generates the answer from the question and the aligned evidence package:

\begin{equation}
A_Q=
\operatorname{Reader}
\left(
Q,\operatorname{Serialize}(E_Q,Z_Q)
\right),
\label{eq:bide_answer}
\end{equation}

Here, $\operatorname{Serialize}(E_Q,Z_Q)$ denotes the formal Reader interface: evidence is presented together with its demand-relevant organization. In the measured compact execution, this organization is carried by the actual Reader inputs \texttt{records}, \texttt{dialogue\_blocks}, and \texttt{reading\_guide}; a separate four-state $Z_Q$ object is not serialized. The explicit ALIGN/VERIFY pilot is a distinct configuration and is excluded from the Full results.

%% file: sec/4_experiments.tex
\section{Experiments}
\label{sec:experiments}

\subsection{Experimental Setup}

We evaluate BiDE-Mem on LoCoMo (1,540 answerable questions, Categories 1--4) and LongMemEval (500 questions). On LoCoMo, we compare against MemoryBank, ReadAgent~\cite{lee2024readagent}, MIRIX, Mem0, EverMemOS~\cite{hu2026evermemos}, and Mnemis~\cite{tang2026mnemis}, using GPT-5.4-mini as the shared Reader. On LongMemEval, we compare with reported baseline results; BiDE-Mem uses Gemma 4 31B and Llama 3.1 8B Instruct for memory processing and GPT-5.4-mini for answering and evaluation. Both benchmarks follow their respective evaluation protocols.

We evaluate seven systems with a shared Qwen3.8-27B Reader on LoCoMo and LongMemEval-S, using Qwen and GPT Judges. System-specific memory sources, retrieval procedures, and prompts remain distinct, making this a shared-Reader rather than identical-memory comparison. Detailed configurations, evaluation protocols, and implementation details are provided in Appendix~\ref{app:experimental}.

\subsection{Main Results}

\paragraph{LoCoMo Results.}

Table~\ref{tab:locomo_main} summarizes the performance comparison on LoCoMo.

\begin{table}[ht]
\centering
\caption{Performance comparison on LoCoMo answerable questions (Cat1--Cat4). Input tokens are averaged per question~(original answer-stage accounting).}
\label{tab:locomo_main}
\small
\resizebox{\linewidth}{!}{%
\begin{tabular}{lcccccc}
\toprule
Method
& Overall
& Multi-hop
& Temporal
& Open-domain
& Single-hop
& Input Tokens \\
\midrule
MemoryBank
&29.61
&26.24
&28.04
&28.13
&31.51
&708.76
\\
ReadAgent
&74.48
&71.99
&60.44
&41.67
&84.42
&4117.46
\\
Mem0
&80.39
&68.79
&78.19
&58.33
&87.63
&3037.61
\\
MIRIX
&83.51
&75.18
&83.18
&55.21
&89.66
&3182.15
\\
EverOS
&80.19
&75.89
&78.50
&64.58
&84.07
&2864.51
\\
Mnemis
&85.26
&69.15
&82.87
&60.42
&94.41
&13370.13
\\
\midrule
BiDE-Mem Top10
&90.13
&84.75
&88.79
&66.67
&95.12
&1334.91
\\
BiDE-Mem Top20
&\textbf{91.49}
&\textbf{87.59}
&\textbf{90.34}
&\textbf{67.71}
&\textbf{95.96}
&2152.04
\\
\bottomrule
\end{tabular}%
}
\end{table}

BiDE-Mem Top20 achieves 91.49\% on LoCoMo, outperforming Mnemis (85.26\%) by 6.23 points. On multi-hop questions it reaches 87.59\%, compared with the strongest baseline result of 75.89\%; Top10 already achieves 90.13\% with a smaller Reader input budget.

\paragraph{LongMemEval Results.}

Table~\ref{tab:longmemeval_main} reports the results on LongMemEval. Baseline numbers correspond to the reported results of the respective systems under their original evaluation settings.

\begin{table}[ht]
\centering
\caption{Performance comparison on LongMemEval.}
\label{tab:longmemeval_main}
\small
\resizebox{\linewidth}{!}{%
\begin{tabular}{lccccccc}
\toprule
Method
& SS-User
& Multi-Session
& SS-Pref.
& Temporal
& Update
& SS-Assist.
& Overall\\
\midrule
MemoryBank
&17.86&42.86&79.17&16.67&19.23&43.33&32.67\\
ReadAgent
&12.50&67.86&8.33&33.33&15.38&53.33&27.72\\
MIRIX
&72.85&30.07&53.33&25.56&52.56&63.63&43.49\\
Mem0
&94.30&66.90&86.70&75.90&87.20&96.40&80.80\\
EverMemOS
&100.00&78.50&96.70&71.20&87.20&78.60&82.00\\
Mnemis
&98.60&86.50&\textbf{100.00}&86.50&93.60&\textbf{100.00}&91.60\\
\midrule
BiDE-Mem
&98.57&\textbf{87.97}&90.00&\textbf{87.22}&\textbf{94.87}&\textbf{100.00}&\textbf{91.80}\\
\bottomrule
\end{tabular}%
}
\end{table}

BiDE-Mem reaches 91.80\% on LongMemEval, compared with the published Mnemis result of 91.60\%. Because the published baselines use different model configurations, Table~\ref{tab:longmemeval_main} provides context rather than a controlled comparison; the shared-Reader results follow.

\paragraph{Unified Qwen Reader Evaluation.}
Table~\ref{tab:qwen_detailed} reports the Qwen3.8-27B Reader results. BiDE-Mem reaches 86.43/89.94\% on LoCoMo and 84.60/88.40\% on LongMemEval-S (Qwen/GPT Judges), outperforming the strongest overall baseline Mnemis by 2.01/2.40 and 6.20/4.00 percentage points, respectively. On LongMemEval-S Temporal questions, it leads Mnemis by 15.04/12.78 points. All seven systems use the shared Reader family, while retaining their memory and retrieval procedures.

\begin{table}[!htbp]
\centering

\caption{Unified Qwen3.8-27B Reader results (accuracy, \%). Entries are Qwen/GPT Judge scores on the same answers; bold marks the best result for each Judge (ties included).}
\label{tab:qwen_detailed}
\scriptsize
\setlength{\tabcolsep}{3pt}
\textbf{(a) LoCoMo (Cat1--Cat4, 1,540 questions)}\\[2pt]
\resizebox{0.97\linewidth}{!}{%
\begin{tabular}{lccccc}
\toprule
Method & Overall & Multi-hop & Temporal & Open-domain & Single-hop \\
\midrule
MemoryBank & 46.30/49.09 & 30.85/34.04 & 41.74/42.68 & 19.79/22.92 & 56.24/59.57 \\
ReadAgent & 71.82/75.97 & 69.15/74.82 & 53.89/60.75 & 54.17/57.29 & 81.57/84.30 \\
Mem0 & 65.52/72.73 & 56.38/68.79 & 65.73/72.59 & 46.88/56.25 & 70.63/75.98 \\
MIRIX & 70.91/75.84 & 71.99/75.53 & 66.67/71.03 & 53.13/64.58 & 74.20/79.07 \\
EverMemOS & 82.92/86.43 & \textbf{82.98}/82.27 & 81.31/86.60 & \textbf{64.58}/\textbf{69.79} & 85.61/89.66 \\
Mnemis & 84.42/87.53 & 71.63/76.60 & 80.69/84.74 & 55.21/63.54 & 93.46/95.01 \\
\midrule
BiDE-Mem & \textbf{86.43}/\textbf{89.94} & 77.66/\textbf{84.04} & \textbf{82.55}/\textbf{87.85} & 59.38/64.58 & \textbf{93.94}/\textbf{95.60} \\
\bottomrule
\end{tabular}%
}

\vspace{0.25em}
\textbf{(b) LongMemEval-S (500 questions)}\\[2pt]
\setlength{\tabcolsep}{2.5pt}
\resizebox{0.97\linewidth}{!}{%
\begin{tabular}{lccccccc}
\toprule
Method & Overall & User & Assistant & Preference & Multi-session & Update & Temporal \\
\midrule
MemoryBank & 39.60/40.80 & 82.86/84.29 & 91.07/92.86 & 16.67/26.67 & 21.05/21.05 & 46.15/47.44 & 15.04/15.04 \\
ReadAgent & 46.20/51.00 & 85.71/85.71 & 82.14/85.71 & 13.33/36.67 & 27.82/33.83 & 58.97/64.10 & 28.57/30.83 \\
Mem0 & 69.40/70.40 & 92.86/94.29 & 73.21/73.21 & 73.33/73.33 & 63.91/65.41 & 85.90/85.90 & 50.38/51.88 \\
MIRIX & 50.80/57.80 & 71.43/80.00 & 46.43/50.00 & 30.00/43.33 & 50.38/54.14 & 66.67/76.92 & 37.59/45.11 \\
EverMemOS & 77.40/80.00 & 84.29/85.71 & 75.00/76.79 & \textbf{83.33}/66.67 & \textbf{75.19}/77.44 & 85.90/\textbf{97.44} & 70.68/73.68 \\
Mnemis & 78.40/84.40 & 92.86/98.57 & \textbf{100.00}/\textbf{100.00} & 53.33/73.33 & 72.18/\textbf{78.20} & \textbf{88.46}/94.87 & 67.67/72.93 \\
\midrule
BiDE-Mem & \textbf{84.60}/\textbf{88.40} & \textbf{97.14}/\textbf{100.00} & 98.21/\textbf{100.00} & 80.00/\textbf{83.33} & 72.93/77.44 & \textbf{88.46}/94.87 & \textbf{82.71}/\textbf{85.71} \\
\bottomrule
\end{tabular}%
}
\end{table}

\subsection{Ablation Study}

We analyze BiDE-Mem under the unified LoCoMo ablation protocol: the same 1,540 questions, GPT-5.4-mini Reader, evaluation procedure, and Top20 evidence budget (Table~\ref{tab:ablation}). The Standard Query Decomposition configuration obtains 87.60\% compared with 91.49\% for Full; removing demand-conditioned retrieval gives 86.69\%, and removing explicit demands gives 78.70\%. Retrieval/reranking and Reader-input field comparisons show the roles of candidate selection and evidence presentation. Appendix~\ref{app:ablation_configuration} specifies the changed components and the archived control correspondence.

\begin{table}[!htbp]
\centering

\caption{LoCoMo component comparisons under the shared GPT-5.4-mini / Top20 protocol (accuracy, \%). $\Delta$ denotes the difference from the submitted Full reference (91.49\%).}
\label{tab:ablation}
\scriptsize
\setlength{\tabcolsep}{2.3pt}
\begin{minipage}[t]{0.485\linewidth}
\centering
\textbf{(a) Demands and organization}\\[2pt]
\resizebox{\linewidth}{!}{%
\begin{tabular}{@{}lrr@{}}
\toprule
Variant & Acc. & $\Delta$ \\
\midrule
Full BiDE-Mem & \textbf{91.49} & -- \\
Standard Query Decomposition & 87.60 & $-3.89$ \\
w/o Demand-conditioned retrieval & 86.69 & $-4.80$ \\
w/o Explicit demands & 78.70 & $-12.79$ \\
w/o Structured memory org. & 86.04 & $-5.45$ \\
w/o Retrieval \& reranking & 68.83 & $-22.66$ \\
\bottomrule
\end{tabular}%
}
\end{minipage}\hfill
\begin{minipage}[t]{0.485\linewidth}
\centering
\textbf{(b) Retrieval and assembly}\\[2pt]
\resizebox{\linewidth}{!}{%
\begin{tabular}{@{}lrr@{}}
\toprule
Variant & Acc. & $\Delta$ \\
\midrule
Single, no reranker & 75.58 & $-15.91$ \\
Single, with reranker & 84.48 & $-7.01$ \\
Multi, no reranker & 84.81 & $-6.68$ \\
Multi, with reranker (Full) & \textbf{91.49} & -- \\
w/o \texttt{dialogue\_blocks} & 85.26 & $-6.23$ \\
w/o \texttt{reading\_guide} & 85.97 & $-5.52$ \\
w/o Both input fields & 84.55 & $-6.94$ \\
\bottomrule
\end{tabular}%
}
\end{minipage}
\end{table}

The retrieval comparison yields 75.58\%/84.48\% for single-channel retrieval without/with reranking and 84.81\%/91.49\% for multichannel retrieval. The single-channel reranking effect is verified from paired question-level outcomes (+137/1,540 questions; +8.90 points). Removing both Reader-facing organization fields yields 84.55\%; the corresponding saved assembly control yields 86.49\% (30-question difference). The submitted Full result is retained as the common headline reference, while the Appendix distinguishes the saved control for each intervention.

\subsection{Demand--Evidence Alignment Analysis}
\label{sec:demand_alignment}

We assess demand--evidence coverage offline on all 1,540 LoCoMo questions (2,517 annotated demands) under Top20, distinct from the 250-question MemPath diagnostic subset.

\begin{figure}[ht]
    \centering

    \begin{subfigure}{0.48\linewidth}
        \centering
        \includegraphics[width=\linewidth]{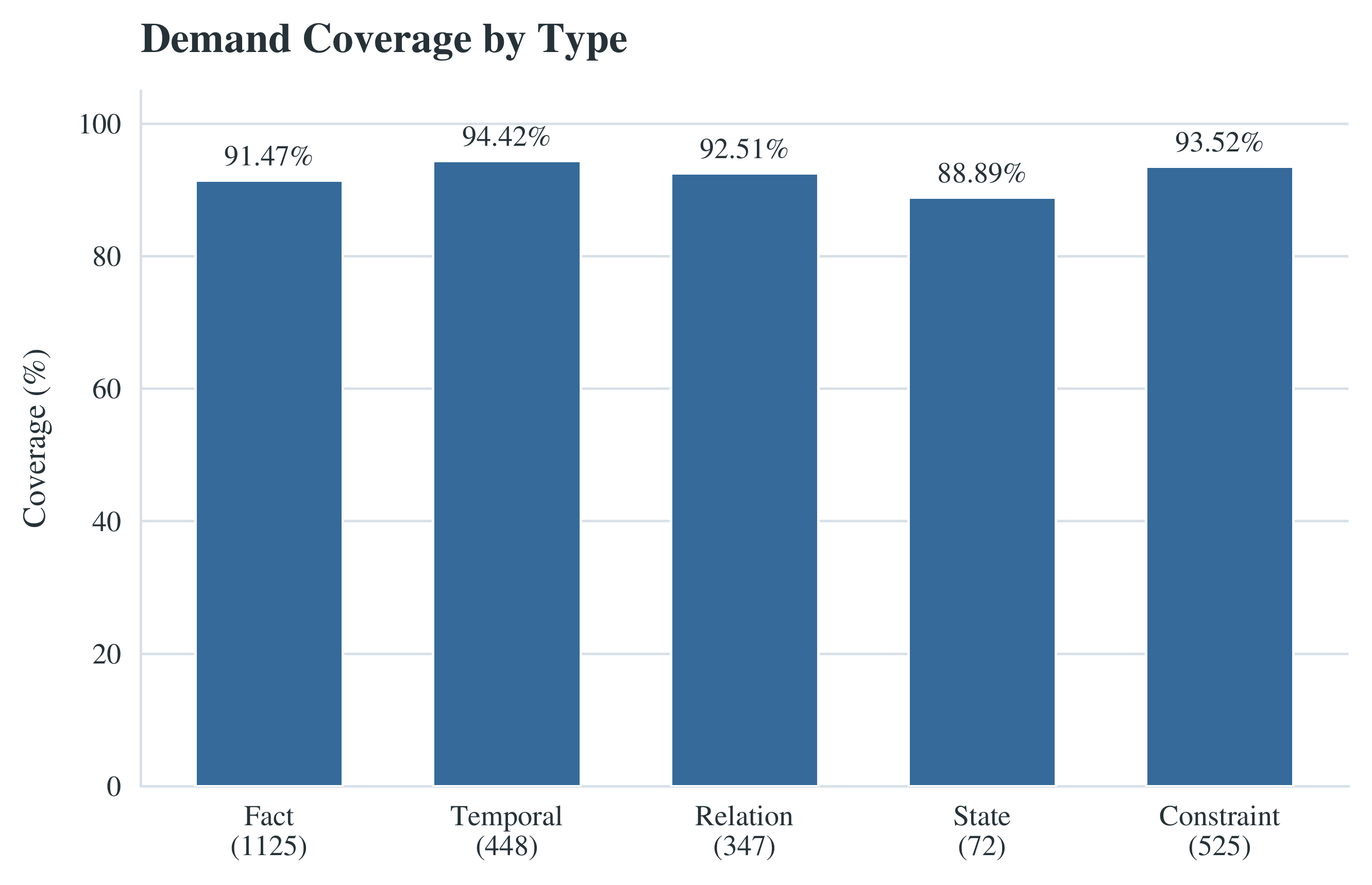}
        \caption{Demand coverage by type.}
        \label{fig:demand_coverage}
    \end{subfigure}
    \hfill
    \begin{subfigure}{0.48\linewidth}
        \centering
        \includegraphics[width=\linewidth]{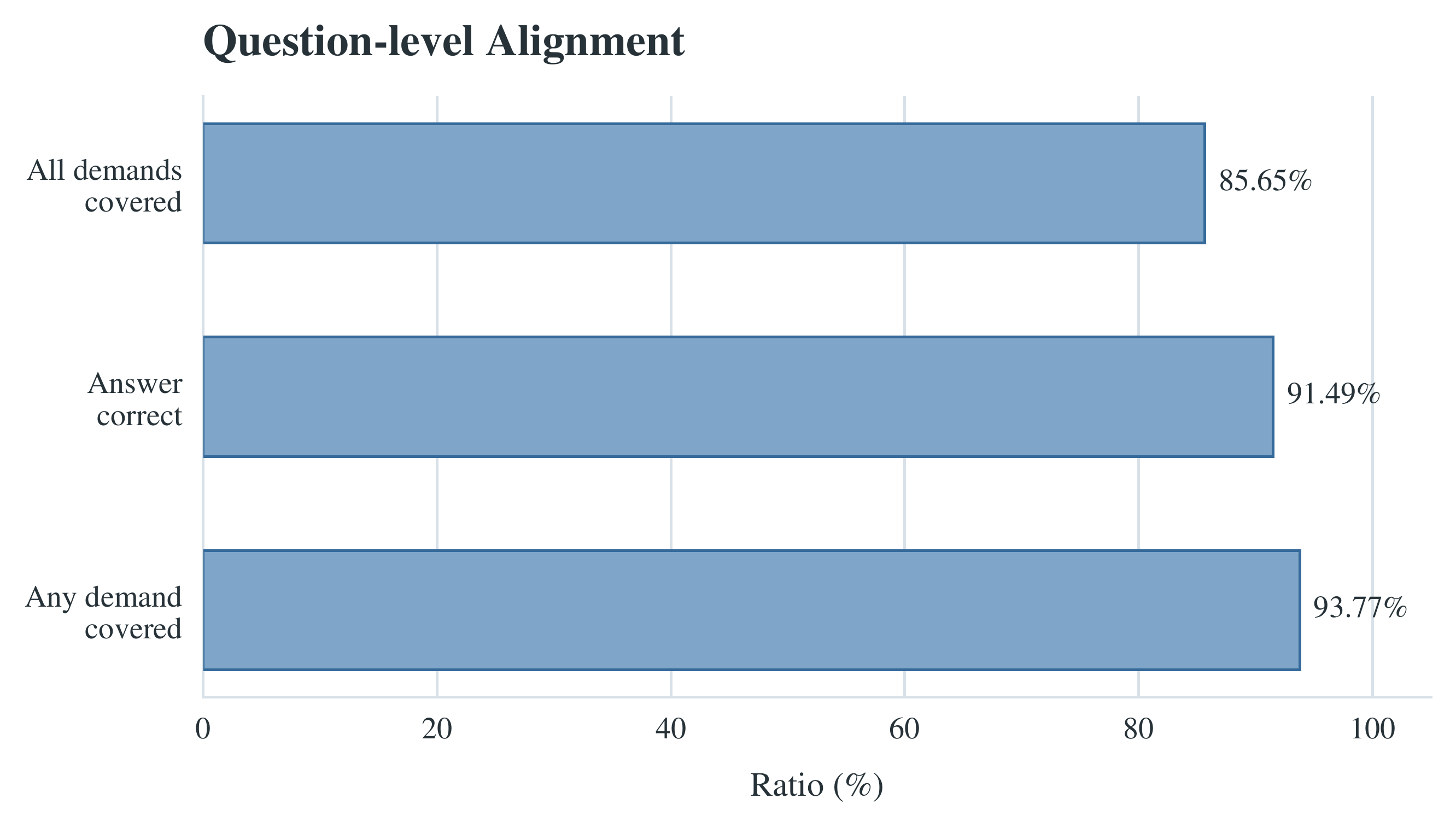}
        \caption{Question-level alignment.}
        \label{fig:question_alignment}
    \end{subfigure}

    \caption{Analysis of demand--evidence alignment in BiDE-Mem.}
    \label{fig:demand_alignment}
\end{figure}

Figure~\ref{fig:demand_coverage} reports 92.49\% annotated-demand coverage, with State demands lowest at 88.89\%. Question-level any-demand and complete-demand coverage are 93.77\% and 85.65\%, respectively (Figure~\ref{fig:question_alignment}). Of 131 incorrect answers, 95 are marked demand-incomplete and 36 demand-complete under the offline metric. These counts correspond to 1,283/1,319 correct answers among complete-demand questions (97.27\%) and 126/221 among incomplete-demand questions (57.01\%). The 221 incomplete questions comprise 125 with partial demand coverage and 96 with no covered demand. This is a descriptive association under the annotation-based coverage criterion, not a causal effect; an incomplete annotation-level coverage result does not rule out alternative or unannotated evidence. Gold-demand annotations were used for offline evaluation, not as online QueryFrame states.

\subsection{Online Efficiency}
\label{sec:efficiency}
Table~\ref{tab:online_cost} reports saved LoCoMo stage-level measurements. In the measured Qwen compact path, BiDE-Mem has two logical generative operations per question (QueryFrame and Reader), with 2,740.65 input and 349.27 output tokens per question on the specified accounting basis. Reranking adds a separately measured 4.35\,s/question of batch-amortized local computation. These records do not establish a full end-to-end latency.

\begin{table}[!htbp]
\centering

\caption{Saved LoCoMo stage measurements from method-specific Qwen answer paths. Token counts exclude upstream processing. Timings are not comparable end-to-end latencies.}
\label{tab:online_cost}
\scriptsize
\setlength{\tabcolsep}{3.8pt}
\resizebox{\linewidth}{!}{%
\begin{tabular}{@{}lrrrrrrr@{}}
\toprule
Metric & BiDE-Mem & Mnemis & EverMemOS & Mem0 & MIRIX & ReadAgent & MemoryBank \\
\midrule
Answer-stage input (tok./Q) & 2,293 & 12,539 & 6,208 & 3,049 & 24,515 & 5,171 & 770 \\
Answer-stage output (tok./Q) & 59 & 110 & 956 & 7 & 170 & 30 & 16 \\
Retrieval (s/Q) & $\approx$0.59$^{\dagger}$ & 16.88 & 40.33 & 0.22 & 6.57 & 8.10 & 0.20 \\
Answer stage (s/Q) & 5.24 & 23.98 & 137.15 & 3.85 & 45.25 & 15.03 & 2.86 \\
\bottomrule
\end{tabular}%
}
\vspace{1pt}
\par\raggedright{\scriptsize $^{\dagger}$ Five retrieval stages only; timings and Reader token counts come from stage-specific records and are not a matched end-to-end comparison.}
\end{table}

%% file: sec/6_discussion_limitations.tex
\section{Discussion and Limitations}
\label{sec:limitations}

\paragraph{Interpretation of the diagnostic subset.}
The cross-system MemPath analysis uses a reviewed subset of 250 LoCoMo questions with 425 Question Demands, whereas the full BiDE-Mem demand analysis covers 1,540 questions and 2,517 demands. The diagnostic subset is used to inspect system-level failure mechanisms; its error and recovery rates should not be interpreted as population estimates for the full LoCoMo benchmark. The current experimental record establishes the subset size and review scope but does not provide sufficient sampling provenance to claim probability sampling or a representative sample design.

\paragraph{Source coverage and semantic sufficiency.}
MemPath's source-grounded coverage measures whether annotated supporting information is retained in the observed evidence context, using source identifiers for source-preserving records and semantic preservation checks for transformed representations. This is an annotation-relative coverage measure, not a complete proof of semantic sufficiency. A demand may be answerable through evidence not included in the annotated support set, and complete annotated-source coverage does not guarantee that the Reader will use the evidence correctly. The full demand inventory and the manually reviewed diagnostic subset also have different review scopes, as detailed in Appendix~\ref{app:dataset}.

\paragraph{Oracle intervention.}
The intervention results are reported over 869 valid system--question cases per setting from 915 initial errors. The retained aggregate record does not establish identical case membership or complete per-case exclusion reasons across the two intervention settings. We therefore report the recovery rates descriptively and do not interpret their difference as a paired causal effect or claim statistical significance from the aggregate counts alone.

\paragraph{Formal support mapping and evaluated execution.}
The four-state $Z_Q$ definition specifies a semantic interface for describing how evidence supports each demand. The reported compact Full execution instead passes \texttt{records}, \texttt{dialogue\_blocks}, and \texttt{reading\_guide} to the Reader; it does not independently serialize a four-state $Z_Q$ object or invoke the separate ALIGN/VERIFY pilot. Accordingly, the reported results support the evaluated demand-guided retrieval and evidence-organization pipeline, not a claim that an independent four-state verifier was evaluated as part of Full.

\paragraph{Cross-system comparability.}
The evaluated memory systems retain system-specific memory representations, retrieval procedures, adapters, and evidence interfaces. The shared-Reader supplementary experiment controls the answer-generation model family and judging protocols, but it does not make the underlying memory inputs or retrieval procedures identical. Published LongMemEval baseline scores likewise come from their original configurations and should be read as contextual reference points rather than fully controlled comparisons.

\paragraph{Efficiency measurement.}
The reported timing and token measurements are stage-specific and use prepared memory or query inputs in several settings. Some retrieval timings cover only a subset of the implemented stages. They therefore characterize the recorded components, not a matched end-to-end latency comparison across systems or benchmarks.

%% file: sec/6_conclusion.tex
\section{Conclusion}
\label{sec:conclusion}

This work studies long-term memory from the perspective of question demands and evidence sufficiency. We introduce MemPath to diagnose demand--evidence misalignment by tracing information accessibility, demand coverage, and evidence effectiveness across the memory pathway. Guided by this diagnosis, BiDE-Mem integrates demand representation, demand-conditioned retrieval, evidence organization, and demand--evidence support mapping into a unified pipeline. Experiments on LoCoMo and LongMemEval demonstrate strong performance in the original evaluations and in supplementary shared-Reader comparisons, with gains varying across categories and judging protocols. Overall, our results show that accessing relevant memories does not necessarily yield sufficient evidence for answering, motivating a shift from retrieval-oriented memory optimization toward demand-aware evidence construction.

\subsection*{AI Use Statement}

In this work, we used generative AI models as components of the proposed method, including memory construction, demand-conditioned query representation, answer generation, and answer evaluation. We also used generative AI tools for assisting manuscript refinement and improving presentation clarity.

We used recorded experimental configurations and reviewed the resulting analyses and manuscript claims. The authors take responsibility for the final content, including the reported measurement scope and limitations.

\subsection*{Reproducibility Statement}

We provide detailed descriptions of the datasets, evaluation protocols, model configurations, baseline settings, and implementation details in the main paper and Appendix. Our experiments use the publicly available LoCoMo and LongMemEval benchmarks and follow their corresponding evaluation protocols. The Appendix further documents the MemPath diagnostic procedure, BiDE-Mem configuration, demand annotations, retrieval and evidence settings, and prompt contracts. The public release should include versioned source code, evaluation scripts, and experiment configurations corresponding to the reported results; the repository URL will be added to the public version.

%% file: sec/7_Ap.tex
\appendix
\section{Experimental Details}
\label{app:experimental}
\subsection{Dataset and Evaluation Setup}
\label{app:dataset}

We evaluate BiDE-Mem on LoCoMo and LongMemEval. LoCoMo is used for both the main evaluation and the MemPath diagnostic analysis. We focus on Categories 1--4, resulting in 1,540 answerable questions. LongMemEval is used as an additional benchmark following its official 500-question evaluation setting and task definitions.

For LoCoMo, we use two demand annotation sets for different analyses. The \textbf{MemPath diagnostic subset} contains 250 questions and 425 manually reviewed Question Demands and is used for the cross-system diagnosis in Section~\ref{sec:mempath}, including source-grounded demand completeness and evidence intervention. The \textbf{full demand analysis} covers all 1,540 LoCoMo questions and contains 2,517 annotated Question Demands. It is used for the full-scale demand--evidence alignment analysis of BiDE-Mem in Section~\ref{sec:demand_alignment}. The 425-demand subset supports detailed diagnosis across heterogeneous memory systems, while the 2,517-demand inventory supports complete analysis of BiDE-Mem over the full LoCoMo evaluation set.

\begin{table}[ht]
\centering
\caption{Dataset statistics and usage in our experiments.}
\label{tab:dataset_statistics}
\small
\begin{tabularx}{\linewidth}{>{\raggedright\arraybackslash}p{0.25\linewidth}>{\raggedright\arraybackslash}p{0.22\linewidth}>{\raggedright\arraybackslash}p{0.17\linewidth}X}
\toprule
Setting & Usage & Scale & Annotation \\
\midrule
LoCoMo & Main evaluation & 1,540 questions & Gold source evidence \\
MemPath diagnostic subset & Six-system diagnosis & 250 questions & 425 reviewed demands \\
Full demand analysis & BiDE-Mem analysis & 1,540 questions & 2,517 demands \\
LongMemEval & Main evaluation & 500 questions & Official evaluation protocol \\
\bottomrule
\end{tabularx}
\end{table}

The LoCoMo evaluation contains 282 multi-hop, 321 temporal, 96 open-domain, and 841 single-hop questions, corresponding to Categories 1--4. Category 5 is excluded. LongMemEval contains 78 knowledge-update, 133 multi-session, 56 single-session-assistant, 30 single-session-preference, 70 single-session-user, and 133 temporal-reasoning questions. Question identifiers define the evaluation sets and link answers to judgments and source records.

Answer availability, judgment availability, and correctness are separate quantities. A saved answer does not imply that evaluation has completed, and an unscored answer is not automatically an incorrect answer. Full-set accuracy uses the benchmark denominator only after each item has a valid judgment or an explicitly documented terminal-failure label. A partial evaluation reports its evaluated denominator and outstanding items separately. The two demand inventories serve offline analysis; the statement that the diagnostic subset is manually reviewed does not establish full manual review of the larger inventory.

\subsection{LongMemEval Configuration}
\label{app:longmemeval_config}

For LongMemEval, BiDE-Mem uses a staged pipeline for memory processing, answer generation, and evaluation, with the model configuration summarized in Table~\ref{tab:longmemeval_config}. This table describes BiDE-Mem in the original LongMemEval evaluation; the published baseline comparison and the supplementary Qwen evaluation have separate protocols.

\begin{table}[ht]
\centering
\caption{LongMemEval model configuration.}
\label{tab:longmemeval_config}
\small
\begin{tabularx}{\linewidth}{>{\raggedright\arraybackslash}p{0.24\linewidth}>{\raggedright\arraybackslash}p{0.19\linewidth}X}
\toprule
Stage & Model & Function \\
\midrule
Preprocessing & Gemma 4 31B & Content parsing and chunk processing \\
Extraction and analysis & Llama 3.1 8B Instruct & L1 mentions, L2 facts, entity relations, and source linking \\
Answer generation & GPT-5.4-mini & Final answer generation \\
Judge & GPT-5.4-mini & Evaluation with the official judging prompt \\
\bottomrule
\end{tabularx}
\end{table}

Gemma 4 31B is used to parse the interaction history and construct the initial memory representation. Llama 3.1 8B Instruct performs structured extraction and analysis, including L1 mentions, L2 factual units, entity--relation information, and source linking. GPT-5.4-mini generates the final answer from the constructed evidence package and is also used with the official LongMemEval judging prompt for evaluation.

For the original LongMemEval comparison, baseline results are drawn from the corresponding published evaluations under their original model and evaluation settings. These baseline scores are not results of our supplementary local reruns.

\paragraph{Supplementary evaluation.}

In the supplementary evaluation, all generative LLM operations executed by the seven adapted systems use Qwen3.8-27B, including memory processing and query/agent operations where applicable, and final answer generation. System-specific memory representations, retrieval procedures, and answer prompts are preserved; the LongMemEval-S Mnemis experiment uses author-provided memory contexts rather than rebuilding them. Qwen3.8-27B and GPT-5.4-mini Judges evaluate the same saved answers using the benchmark judging protocols. This setting is separate from the original LongMemEval model configuration in Table~\ref{tab:longmemeval_config}. Appendices~\ref{app:baseline} and~\ref{app:supplementary} describe the evaluated systems and results.

\subsection{Baseline Evaluation}
\label{app:baseline}
\label{app:reproduction_audit}
\label{app:source_reproduction}

For the original GPT LoCoMo comparison using a shared answer model, we evaluate BiDE-Mem against six representative long-term memory systems: Mnemis, MIRIX, Mem0, EverOS, ReadAgent, and MemoryBank.

The evaluated systems adopt different memory representations, retrieval interfaces, and answer-generation procedures. We therefore preserve each system's native memory representation, retrieval procedure, evidence interface, and answer prompt rather than replacing them with a unified memory pipeline. All systems use GPT-5.4-mini as the shared answer-generation model in that original LoCoMo comparison; the supplementary Qwen comparison is specified separately in Appendix~\ref{app:qwen_protocol}. Final answers are evaluated using the same LoCoMo Judge protocol.

For MemPath diagnosis, system-specific adapters record observable intermediate artifacts and map them to the common diagnostic interface without modifying the underlying memory representation, retrieval behavior, evidence interface, or answer prompt.

\begin{table}[ht]
\centering
\caption{Baseline systems and evaluated artifacts on LoCoMo.}
\label{tab:baseline_artifacts}
\begin{tabularx}{\linewidth}{>{\raggedright\arraybackslash}p{0.17\linewidth}>{\raggedright\arraybackslash}p{0.41\linewidth}X}
\toprule
Method & Evaluated Artifact & Retrieval Interface \\
\midrule
Mnemis & Reconstruction v2 & Memory retrieval \\
MIRIX & Tool-aware evaluation trace & Tool-based retrieval \\
Mem0 & Local reconstruction & Memory retrieval \\
EverOS & Recorded evaluation run & Hybrid retrieval \\
ReadAgent & Page hierarchy reconstruction & Page expansion \\
MemoryBank & Retention-based memory evaluation & Entry retrieval \\
\bottomrule
\end{tabularx}
\end{table}

\paragraph{Source-based implementation and shared evaluation.}

We reproduce the core procedures using the original papers and public source code, adapting benchmark inputs and model interfaces. In the supplementary evaluation, Qwen3.8-27B is used for executed generative LLM stages, and Qwen3.8-27B and GPT-5.4-mini Judges score the same saved answers. Memory schemas, retrieval procedures, tool interactions, and answer prompts remain system-specific; BiDE-Mem's candidate and evidence budgets are not imposed on other systems. The results characterize these adapted system configurations rather than exact reproductions of published scores.

BiDE-Mem uses structured memory and demand-guided retrieval; MIRIX retains tool-based memory access; Mem0 uses locally reconstructed entries and memory retrieval; ReadAgent uses page hierarchies and page expansion; and MemoryBank retrieves retained entries. The EverMemOS adapter connects hybrid memory search to the evaluation interface, renders returned episodes as Reader context, and invokes the shared generation model. The original EverOS artifact in Table~\ref{tab:baseline_artifacts} and the supplementary EverMemOS runs have separate execution records.

\paragraph{Mnemis on LongMemEval.}

For LongMemEval-S, Mnemis uses the authors' released evaluation contexts as its memory input; for LoCoMo, it uses the locally reconstructed memory and retrieval interface. Supplementary answers use the Qwen3.8-27B Reader and are evaluated with the stated Qwen3.8-27B and GPT-5.4-mini Judge protocols.

The two Judges evaluate the same saved answer for each question wherever paired judgments are available. The final benchmark results are reported in Appendix~\ref{app:supplementary}.

\subsection{BiDE-Mem Configuration on LoCoMo}
\label{app:configuration}

For the original GPT LoCoMo experiments, BiDE-Mem uses GPT-5.4-mini for LLM-based memory construction, demand-conditioned query representation, and final answer generation. Retrieval and evidence construction combine multiple retrieval channels, reranking, and budget-controlled evidence selection. Final answers are evaluated using the same Judge protocol as the baseline systems.

The main implementation configurations are summarized in Table~\ref{tab:implementation_configuration}.

\begin{table}[ht]
\centering
\caption{Implementation configuration of BiDE-Mem in the original GPT LoCoMo setting.}
\label{tab:implementation_configuration}
\begin{tabular}{ll}
\toprule
Component & Configuration \\
\midrule
LLM backbone & GPT-5.4-mini \\
Dense retriever & BGE-M3 \\
Sparse retriever & SPLADE \\
Reranker & Qwen3-8B reranker \\
Candidate pool & 128 candidates \\
Evidence budget & Top10 / Top20 \\
Reader & GPT-5.4-mini \\
Judge & GPT-5.4-mini \\
\bottomrule
\end{tabular}
\end{table}

The original GPT LoCoMo configuration specifies temperature 0 for its LLM-based components. Candidates are collected from multiple retrieval channels, merged, deduplicated, reranked, and selected under the specified evidence budget. The Top10 and Top20 configurations differ only in the final evidence budget unless otherwise stated.

The local computation environment uses an NVIDIA A100 80GB PCIe GPU with CUDA 12.2 and NVIDIA driver 535.309.01. This hardware description concerns local computation and does not identify the hardware serving remote model APIs.

The six fusion families are listed in Table~\ref{tab:retrieval_channels}. Dense and lexical rankings also supply intermediate retrieval signals, including seeds for occurrence expansion; their presence as preprocessing stages does not add two independent families to that table. Appendix~\ref{app:evidence_construction} specifies family-level aggregation and candidate selection. The 128-candidate retrieval budget and the reranker's token budget constrain different objects: the former limits candidate records, whereas the latter limits the text supplied for each candidate score. Long-input windowing and score aggregation therefore belong to the run configuration rather than being implied by Top-$K$ alone.

\section{MemPath Diagnostic Framework Details}
\label{app:mempath}
\subsection{System Adaptation to the Diagnostic Interface}
\label{app:system_adaptation}

MemPath provides a unified diagnostic interface for analyzing how heterogeneous long-term memory systems transform question-relevant historical information into answer-supporting evidence. Following the formulation in the main text, the operational diagnostic interface is
\begin{equation}
Q\rightarrow D(Q)\rightarrow C_Q\rightarrow (E_Q,Z_Q)\rightarrow A_Q ,
\label{eq:mempath_diagnostic_interface}
\end{equation}
where $D(Q)$ denotes the Question Demands used as the common diagnostic reference, $C_Q$ denotes the candidate information exposed by the system, $E_Q$ denotes the final evidence context available for answering, $Z_Q$ records the relationship between the observed evidence and the annotated demands, and $A_Q$ denotes the system answer.

The internal memory construction and organization mechanisms preceding $C_Q$ remain system-specific. Existing long-term memory systems expose different memory representations, retrieval mechanisms, and observable intermediate states. MemPath therefore does not require identical architectures. Instead, we construct system-specific adapters that map available artifacts into the diagnostic interface while preserving the original behavior of each system.

\begin{table}[ht]
\centering
\caption{System-specific artifacts mapped to the MemPath diagnostic interface.}
\label{tab:mempath_adapters}
\small
\begin{tabularx}{\linewidth}{llX}
\toprule
System & Memory Artifact & Retrieval Artifact \\
\midrule
Mnemis & Memory units and event structures & Retrieved memory records \\
MIRIX & Episodic memory and tool traces & Tool-based retrieval outputs \\
Mem0 & Stored memory entries & Retrieved memory records \\
EverOS & Available memory traces & Retrieved memory context \\
ReadAgent & Page-level memory hierarchy & Expanded reading pages \\
MemoryBank & Retained memory entries & Retrieved memory items \\
\bottomrule
\end{tabularx}
\end{table}

The adapters only collect observable intermediate artifacts and associated metadata. They do not modify the memory construction, retrieval, Reader, answer prompt, or answer-generation procedure of the evaluated systems.

The diagnostic $Z_Q$ is an offline description of the relationship between observed evidence and annotated demands. It does not imply that the evaluated baseline natively computes or consumes BiDE-Mem's alignment representation. Native memory objects, retrieved records, actual Reader inputs, and generated answers are distinguished from the demand labels and support annotations introduced for diagnosis.

Observability is stage-specific. A missing trace or an unavailable source mapping is an unobserved measurement, whereas a complete trace with no matching support is an observed absence. Source coverage requires a defined source mapping and reports its eligible denominator. A source recovered from a different execution cannot establish that the evaluated Reader saw that source.

\subsection{Question Demand Annotation}
\label{app:demand_annotation}

MemPath defines a Question Demand as an atomic answer requirement that the available evidence must satisfy. To support demand-level diagnosis, we construct a manually reviewed demand annotation set from 250 LoCoMo questions, containing 425 Question Demands.

Each question is decomposed into one or more atomic demands according to the information required by the question and reference answer. Each demand is further associated with supporting source evidence. The annotation procedure consists of:
\begin{enumerate}
\item identifying answer-bearing requirements from the question and reference answer;
\item decomposing these requirements into atomic Question Demands;
\item associating each demand with the source evidence required to support it.
\end{enumerate}
Each demand is represented by a structured record containing its semantic requirement and evidence dependencies.

\begin{table}[ht]
\centering
\caption{Question Demand annotation schema.}
\label{tab:demand_schema}
\begin{tabularx}{\linewidth}{lX}
\toprule
Field & Description \\
\midrule
Demand type & Fact, Relation, Temporal, State, or Constraint \\
Target entity/event & Entity or event involved in the requirement \\
Relation requirement & Required relationship or dependency \\
Temporal condition & Required time scope or ordering constraint \\
State condition & Required state or transition information \\
Supporting sources & Source IDs providing required evidence \\
\bottomrule
\end{tabularx}
\end{table}

The 425-demand annotation set is used only for the cross-system MemPath diagnosis. Separately, the full-scale BiDE-Mem analysis uses the 2,517-demand inventory described in Appendix~\ref{app:dataset}. These evaluation annotations are not provided to BiDE-Mem during memory construction, retrieval, evidence construction, or answer generation.

\paragraph{LLM-assisted annotation and human cross-review.}

Initial demand annotations were prepared using an LLM with questions, reference answers, and supporting sources. Two PhD researchers participated in cross-review for requirement completeness, demand decomposition, type consistency, and source grounding. The reviewed diagnostic inventory contains 250 questions and 425 demands; the separate full-scale analysis contains 1,540 questions and 2,517 annotated demands. These are inventory sizes, not per-reviewer workloads. No independent initial labeling record or agreement coefficient is available; the larger inventory is not claimed to have undergone full dual independent review.

Per-reviewer workloads and item-level adjudication counts are not available in the retained summary and therefore are not estimated.

\paragraph{Interpreting the annotation fields.}

A Fact demand concerns an attribute or value; a Relation demand concerns a dependency between entities or events; a Temporal demand concerns a date, duration, or ordering; a State demand concerns a condition or transition; and a Constraint demand restricts which candidate answers are admissible. An atomic requirement can retain several qualifiers. For example, a temporal requirement about a preference change still retains the relevant person and activity. Splitting these qualifiers into unrelated demands would lose the dependency needed to answer the question.

\paragraph{Illustrative annotation.}

For ``When did Alice stop enjoying jazz?'', the requested value is the time of a particular preference transition. The statement ``Alice stopped enjoying jazz after moving to Boston'' supports a relative ordering with respect to the move, but does not supply a calendar date. Its source identifier anchors the annotation; the date must remain unspecified unless another source provides it. This example illustrates the schema and does not add an item to either annotated inventory.

The reference answer and supporting sources belong to offline annotation. Online QueryFrame construction receives the question without those annotation inputs. The two uses of a demand representation consequently have different information access: an annotated demand can identify known support sources, whereas an online query specifies what support to seek.

\subsection{Diagnostic Evaluation Protocol}
\label{app:diagnostic_protocol}
\label{app:failure_policy}
\label{app:diagnostic_counts}

MemPath evaluates memory systems from three complementary perspectives: source accessibility, demand completeness, and evidence intervention.

\paragraph{Source-level accessibility.}
Source-level analysis examines whether annotated supporting information successfully propagates into the observable retrieval and answer contexts of each system. Given a set of annotated supporting sources, we measure whether these sources appear in the retrieved candidate set and whether they remain available in the final context used for answering. These measurements correspond to the retrieval and prompt coverage statistics reported in the main text.

\paragraph{Demand completeness.}
Demand-level analysis evaluates whether the observed evidence retains the supporting information annotated for each Question Demand. Let $A_i$ denote the annotated supporting source set for demand $d_i$, and let $S_i$ be the set of those sources retained for $d_i$ at the evaluated stage. For uncompressed source-preserving records, source identifiers establish that the original source content remains available. For compressed summaries and structured memory units, we assess whether the retained representation can be linked to the original source and preserves the fact, relation, temporal condition, state, or constraint needed by $d_i$. This demand-indexed source matching handles the case in which one compressed record preserves support for some demands but not others. It retains the same all-support rule for every memory representation; only the source-retention test differs.

The any-anchor criterion considers a demand accessible when at least one supporting source is available:
\begin{equation}
z_i^{\mathrm{any}}=\mathbb{I}[A_i\cap S_i\neq\emptyset].
\label{eq:any_anchor}
\end{equation}
The all-support criterion requires all annotated supporting sources for the demand to be available:
\begin{equation}
z_i^{\mathrm{all}}=\mathbb{I}[A_i\subseteq S_i].
\label{eq:all_support}
\end{equation}

The any-anchor criterion measures availability of at least one annotated supporting source, whereas all-support requires coverage of the entire annotated support set. Both apply the same representation-aware source-retention rule. For a question $Q$, the source-grounded demand coverage is
\begin{equation}
\operatorname{Cov}(E_Q,D(Q))=\frac{1}{|D(Q)|}\sum_{d_i\in D(Q)}z_i^{\mathrm{all}} .
\label{eq:appendix_demand_coverage}
\end{equation}
A question is termed \emph{demand-complete} in the diagnostic plots when all its annotated demand support sets satisfy the all-support criterion, i.e., $\operatorname{Cov}(E_Q,D(Q))=1$. This is the measure shown in Figure~\ref{fig:diagnosis_demands}; it measures evidence coverage rather than whether the Reader ultimately generates the correct answer.

For compressed or transformed evidence, semantic mapping examines the retained demand-relevant content, including entity identity, relation direction, temporal scope, modality, and state changes; retaining a provenance ID without that content does not count as support. This content-sensitive check is part of determining source retention for the corresponding memory representation, not a replacement for the all-support definition. Final answer correctness remains a separate benchmark outcome.

The criterion is defined relative to the annotated source set $A_i$. A different sufficient source set need not be included in $A_i$, so this is a conservative annotation-grounded measure rather than a search over every logically equivalent proof. Empty or unresolved support annotations are excluded from valid all-support judgments rather than counted as complete through the empty-set identity.

\paragraph{Evidence intervention.}
For incorrect answers, MemPath first distinguishes demand-incomplete cases from demand-complete but wrong cases. We then intervene on the evidence while keeping the Reader and answer-generation procedure unchanged. In the \textbf{Evidence-only} setting, the original answer context is replaced with verified supporting evidence. In the \textbf{Combined} setting, missing verified evidence is added to the information already available to the system, and the resulting evidence is reorganized into a coherent answer-ready context before being passed to the Reader. All intervention outputs are evaluated using the same answer-level Judge protocol.

The corresponding recovery rate for an intervention $j$ is
\begin{equation}
\operatorname{Recovery}(j)=\frac{1}{|\mathcal{F}_j|}\sum_{(s,Q)\in\mathcal{F}_j}\mathbb{I}[A_{s,Q}^{(j)}\text{ is correct}],
\label{eq:intervention_recovery}
\end{equation}
Here, $\mathcal{F}_j$ is the set of valid evaluated system--question pairs drawn from the initially incorrect pairs. The index $s$ identifies the memory system, and $A_{s,Q}^{(j)}$ is its answer after intervention $j$.

\paragraph{Diagnostic counts and intervention validity.}

The diagnostic unit is a system--question pair. Applying six systems to the same 250 questions yields 1,500 possible pairs; the 425 annotated demands are shared question annotations rather than 425 independently sampled demands per system. Within each system, demand-complete questions and incorrect answers are intersecting sets. The number of incorrect answers therefore cannot be relabeled as the number of demand-complete but wrong answers.

The initial-error set and the evaluated intervention set $\mathcal{F}_j$ have distinct roles. The first defines potential recovery cases; the second defines the denominator in Equation~\ref{eq:intervention_recovery}. A paired comparison between Evidence-only and Combined uses the intersection of their valid evaluated system--question identifiers. If the two sets differ, their separate denominators and the paired intersection must remain distinguishable. An aggregate recovery count alone does not identify which original errors were excluded or establish matched intervention conditions.

The aggregate diagnostic ledger contains 915 initially incorrect system--question pairs and 869 valid intervention cases. Evidence-only recovers 404 cases (46.49\%), and Combined recovers 476 (54.78\%). The system-level counts are shown below. The six rows share the same 250 questions; their denominators count system--question pairs.

{
\begin{center}
\small
\resizebox{\linewidth}{!}{%
\begin{tabular}{lrrrrr}
\toprule
System & Complete/250 & Errors & Valid & Evidence-only & Combined \\
\midrule
Mnemis & 192 & 110 & 101 & 38 & 50 \\
MIRIX & 95 & 103 & 93 & 49 & 46 \\
Mem0 & 183 & 150 & 134 & 63 & 67 \\
EverOS & 147 & 158 & 158 & 82 & 107 \\
ReadAgent & 103 & 164 & 157 & 64 & 93 \\
MemoryBank & 34 & 230 & 226 & 108 & 113 \\
\midrule
Total & -- & 915 & 869 & 404 & 476 \\
\bottomrule
\end{tabular}%
}
\end{center}
}

These are reconstructed aggregate counts. The 46-case difference between initial errors and valid cases lacks an item-level exclusion ledger, and the aggregates do not establish matched identities between intervention settings. The recovery percentages are descriptive and cannot justify a paired significance test from these totals alone. ReadAgent's 164 initial errors are not demand-complete errors: its source-complete set contains only 103/250 questions. Source completeness, initial errors, and intervention cases use different statistical units or eligibility criteria.

\section{BiDE-Mem Implementation Details}
\label{app:bide_details}
\subsection{Memory Assertion Construction}
\label{app:memory_assertion}

BiDE-Mem constructs structured memories from raw interactions to support demand-aware retrieval and evidence construction. The memory organization is demand-aware by design rather than reconstructed for each question: the predefined Question Demand framework determines which factual, relational, temporal, state, and provenance information should be preserved so that future question-specific demands can access it reliably.

Rather than storing only compressed summaries, BiDE-Mem preserves fine-grained assertions together with their original provenance. Given an interaction history, the memory construction process transforms raw records into a hierarchical representation:
\begin{equation}
\begin{aligned}
\text{Raw Interactions}&\rightarrow\text{L1 Propositions}\rightarrow\text{L2 Assertions}\\
&\rightarrow\text{L3 Indices}.
\end{aligned}
\label{eq:memory_hierarchy}
\end{equation}
The construction process first identifies clause-level proposition units and then converts them into structured searchable assertions. Each assertion preserves semantic information required for later retrieval and reasoning, including entities, events, relations, values, semantic roles, polarity, temporal information, state information, and provenance.

The detailed construction contract and output format are provided in Appendix~\ref{app:prompt_contract}.

\begin{table}[ht]
\centering
\caption{Hierarchical memory representation in BiDE-Mem.}
\label{tab:memory_hierarchy}
\begin{tabularx}{\linewidth}{llX}
\toprule
Level & Representation & Function \\
\midrule
Raw & Original conversation records & Preserve source information \\
L1 & Clause-level proposition units & Capture atomic statements \\
L2 & Structured assertions & Support retrieval and reasoning \\
L3 & Entity/event/temporal/state indices & Enable structured access \\
\bottomrule
\end{tabularx}
\end{table}

The constructed memory maintains traceability from structured assertions back to the original interaction records. This design preserves the information structures anticipated by the demand framework while avoiding unsupported cross-session summaries and enabling retrieved evidence to be traced back to its original source.

In the executable memory representation, L1 units retain clause identifiers, source spans, resolution status, and links to L2 assertions. L2 records retain an assertion identifier, subject, relation, answer value, roles, modality, polarity, scope, and provenance. Provenance links the assertion to raw records and clauses, with unit links retained when available. Deterministic compilation converts these records into retrieval views and graph links. Schema conversion is distinct from generating the underlying assertion content.

Memory extraction uses historical interaction records rather than the current evaluation question or reference answer. Contextual resolution, when enabled, is a separate memory-processing branch whose input and provenance must remain distinguishable from direct extraction. Its generation calls belong to memory construction. Online query compilation and Reader calls are accounted for separately in Appendix~\ref{app:cost}.

\subsection{QueryFrame Implementation of Question Demands}
\label{app:queryframe}
\label{app:safeguards}

Question Demand, denoted by $D(Q)$ in the main text, is the conceptual representation of the answer requirements associated with a question. In the implementation, these requirements are instantiated through a structured object referred to as \textbf{QueryFrame}. QueryFrame is therefore not a separate methodological representation; it is the implementation-level realization of the Question Demand formulation.

Given a question $Q$, the QueryFrame compiler converts the requirements implied by the question into structured fields used by downstream memory access and evidence construction. Unlike conventional query rewriting approaches that mainly extract semantic topics or retrieval keywords, QueryFrame represents answer-oriented conditions such as relation dependencies, temporal constraints, answer roles, and required evidence structures.

The compiler contract is described in Appendix~\ref{app:queryframe_contract}.

\begin{table}[ht]
\centering
\caption{QueryFrame implementation fields for Question Demands.}
\label{tab:queryframe_fields}
\begin{tabularx}{\linewidth}{lX}
\toprule
Field & Description \\
\midrule
Subject & Target entity or event \\
Observer & Information holder or perspective entity \\
Evaluated subject & Entity being evaluated or described \\
Relation family & Required relation type \\
Relation direction & Direction of dependency \\
Answer role & Expected answer form \\
Temporal constraint & Required temporal condition \\
Location constraint & Required spatial condition \\
Evidence shape & Required evidence structure \\
List protection & Prevent omission of required items \\
\bottomrule
\end{tabularx}
\end{table}

The QueryFrame representation does not contain answers, retrieved memories, reference answers, gold demand annotations, or external knowledge. It only operationalizes the information requirements implied by the input question.

\paragraph{Executable retrieval representation.}

The current retrieval compiler represents question requirements using a query kind; target entities and owner roles; relation, value, event, and temporal queries; polarity and modality; required evidence acts; graph operations; and lexical expansions. These fields instantiate the conceptual roles in Table~\ref{tab:queryframe_fields}, but the runtime schema and that descriptive table are not identical field lists. Relation queries and lexical expansions provide retrieval text, value queries specify the requested answer slot, and graph operations express potential access paths to linked evidence.

QueryFrame represents answer roles, modality, evidence acts, and graph operations beyond ordinary retrieval subquestions. The Standard Query Decomposition comparison uses the \texttt{without\_\allowbreak parsing} query-input configuration under the same retrieval stack and Top20 budget. Its input representation differs from Full, including the \texttt{condition} field; the available inputs do not separately document a generated subquestion list. The comparison therefore measures the specified query-input configuration rather than a parser-only effect. Gold-demand annotations remain exclusive to offline evaluation.

\paragraph{Structural input checks.}

Input preparation checks for missing queries or rankings, insufficient or duplicate selected evidence, and cross-conversation evidence. The compiler's validation and retry procedure is specified in Appendix~\ref{app:queryframe_contract}; schema validity alone does not establish correct interpretation of a question.

\subsection{Demand-guided Retrieval and Evidence Construction}
\label{app:evidence_construction}

Given the QueryFrame implementation of $D(Q)$, BiDE-Mem retrieves candidate memories through multiple complementary retrieval channels. Different channels target lexical, semantic, entity/event, structural, temporal, and typed evidence signals required by different Question Demands.

At the formulation level, demand-guided retrieval is expressed as follows for each demand $d_i$:
\begin{equation}
C_i=\operatorname{Retrieve}(\mathcal{M},d_i),\qquad C_Q=\bigcup_{i=1}^{m}C_i .
\label{eq:appendix_retrieval}
\end{equation}
Candidates returned by the different access paths are merged into a unified candidate pool, deduplicated, reranked, and selected according to the evidence budget.

\begin{table}[ht]
\centering
\caption{Retrieval channels used in BiDE-Mem.}
\label{tab:retrieval_channels}
\begin{tabularx}{\linewidth}{lX}
\toprule
Channel & Retrieval Target \\
\midrule
Clause retrieval & Proposition-level evidence \\
Multivector retrieval & Semantic similarity \\
Occurrence retrieval & Entity/event mentions \\
SPLADE retrieval & Lexical matching \\
Structural retrieval & Relation-dependent evidence \\
Typed-L2 retrieval & Structured assertions \\
\bottomrule
\end{tabularx}
\end{table}

\paragraph{From query fields to retrieval signals.}

The runtime shares query fields across retrieval families rather than assigning one demand type to one family. For example, Typed-L2 combines the question, target entities, relation paraphrases, event anchors, and temporal cues for predicate matching. A separate value view combines the question, value cues, and target entities. Its owner-oriented ranking uses overlap with assertion subjects. Assertion matches are mapped back to raw record identifiers before candidate fusion. Occurrence retrieval starts from dense and lexical rankings and expands linked records through available antecedent, occurrence, confirmation, denial, and proposition-source edges. These operations retrieve potentially useful records; a high retrieval score does not establish that a demand is satisfied.

\paragraph{Family aggregation and fusion.}

In the inspected six-family RRF implementation used by the Qwen pipeline, each lane contributes at most 128 ranked raw records. Clause, occurrence, structural, multivector, and SPLADE lanes are collapsed within their family using each record's best rank. Typed-L2 first combines its internal lanes by reciprocal-rank fusion. The six family rankings are then combined using contributions $1/(20+r_f)$, where $r_f$ is the record's rank in family $f$ and an absent record contributes zero. Candidates are deduplicated by raw identifier; ties are resolved by ascending identifier. The fused pool contains at most 128 available unique records. Reranking orders this pool before selection of the final evidence budget.

The six-family RRF configuration above describes the supplementary Qwen LoCoMo retrieval path. The GPT--Top20 component interventions are reported in Appendix~\ref{app:ablation_configuration}.

After reranking, the Top-$K$ records are reconstructed into compact evidence units using their original information. The final evidence package preserves:
\begin{itemize}
\item source identifier;
\item speaker information;
\item timestamp;
\item original text;
\item optional image caption;
\item retrieval rank;
\item source session and turn order.
\end{itemize}
Redundant retrieval wrappers are removed while the relationships required for reasoning are preserved. Rather than directly concatenating retrieved memories, BiDE-Mem organizes selected records so that evidence associated with the same event remains connected, temporal order and state changes remain visible, and distinct events or temporal versions remain distinguishable. This produces the structured evidence package $E_Q$ whose support relationships are defined in Appendix~\ref{app:evidence_validation}.

\subsection{Evidence--Demand Alignment}
\label{app:evidence_validation}
\label{app:model_audit}

This subsection defines the demand--evidence alignment representation used in the paper. The evidence-to-demand direction complements demand-guided retrieval by describing how the selected evidence supports the requirements of the question. Given the Question Demands $D(Q)=\{d_1,\ldots,d_m\}$ and the constructed evidence package $E_Q=\{e_1,\ldots,e_n\}$, $Z_Q$ denotes their demand-indexed support and consistency information. It associates each demand with relevant evidence, indicates whether the required conditions are supported, and records missing conditions or conflicting evidence. Thus, $E_Q$ supplies the source information, while $Z_Q$ describes its relationship to the answer requirements.

\paragraph{Demand-level support.}

Support is assessed with respect to the complete requirement expressed by a demand, including its entity, relation, temporal, and state conditions. A record about the correct entity may still concern the wrong event, time, or state. Conversely, several evidence units may jointly support a demand even when no single unit is sufficient. The support assessment therefore concerns the selected evidence collectively:
\begin{equation}
S(d_i,E_Q)\in\{\text{satisfied},\text{partial},\text{conflicting},\text{unmet}\}.
\label{eq:demand_alignment_state}
\end{equation}
The collection of these assessments and their evidence links forms $Z_Q$. Table~\ref{tab:evidence_validation} defines the four states used in the paper.

\begin{table}[ht]
\centering
\caption{Demand--evidence alignment states. Each state concerns support for a particular demand within the selected evidence package.}
\label{tab:evidence_validation}
\begin{tabularx}{\linewidth}{lX}
\toprule
State & Description \\
\midrule
Satisfied & The evidence jointly supports the conditions required by the demand. \\
Partial & Some required conditions are supported, while others remain unresolved. \\
Conflicting & Applicable evidence provides incompatible support that remains unresolved. \\
Unmet & Support for the required information is not identified in the selected evidence. \\
\bottomrule
\end{tabularx}
\end{table}

\paragraph{Evidence links and consistency.}

The alignment retains the connection between a demand and the source-grounded evidence supporting it. Entity consistency concerns who or what the evidence describes. Relation consistency concerns the required dependency and its direction, including which entity occupies each role. Temporal consistency concerns the relevant time interval or ordering. State consistency distinguishes conditions such as a plan, a completed event, and a subsequent change. Statements about different times or events are not inherently conflicting; a conflict concerns incompatible statements within the scope applicable to the demand.

For partially supported demands, the supported information remains associated with the demand alongside the unresolved conditions. For conflicting demands, the incompatible evidence remains distinguishable so that one side is not silently treated as sufficient support. An unmet state describes a gap in the selected package, rather than establishing that the information is absent from the entire interaction history. These distinctions preserve useful evidence without treating topical relevance or source presence as complete support.

\paragraph{Connection to answer generation.}

In the main-text formulation, the answer-ready representation consists of the organized evidence and its alignment information, $(E_Q,Z_Q)$:
\begin{equation}
Q\rightarrow D(Q)\rightarrow C_Q\rightarrow(E_Q,Z_Q)\rightarrow A_Q .
\label{eq:appendix_bide_pipeline}
\end{equation}
The formal Reader interface is expressed as

{
\[
A_Q=\operatorname{Reader}\bigl(Q,\operatorname{Serialize}(E_Q,Z_Q)\bigr).
\]
}

In this formulation, $Z_Q$ specifies which evidence bears on each answer requirement and which conditions remain incomplete or inconsistent. Its intended role at the Reader interface is to guide requirement coverage and the treatment of uncertainty while preserving the original evidence as the factual basis for answering. This equation defines the representation and interface; it does not assert that every evaluated execution serializes a separate four-state object or invokes an independent alignment module. The observed execution forms are described below. Alignment adds no external facts and does not independently supply an answer.

\paragraph{Interpretation in the diagnostic analysis.}

MemPath and BiDE-Mem use distinct evaluation roles for demand support. MemPath checks coverage of annotated supporting sources, applying a semantic preservation check where a compressed or transformed memory replaces the original source text. The formal $Z_Q$ relation instead describes the state of each requirement within the assembled evidence package. Complete source-grounded coverage does not guarantee correct Reader utilization. The post-retrieval field ablations test organization and guidance inputs, not an independently generated four-state verifier.

\paragraph{Observed execution forms.}

The retained explicit alignment implementation executes ALIGN, VERIFY, evidence assembly, and Reader generation, supplying demand-indexed support records before answering. Its three serialized labels represent partial support through supporting units and missing conditions under \texttt{unmet}. This implementation is a post-reranking pilot; its extra calls are not established as part of the reported Full configuration. The four-state formulation above describes the semantic interface.

The 1,540 retained GPT LoCoMo answer calls instead contain \texttt{records}, \texttt{dialogue\_blocks}, and \texttt{reading\_guide}, with parsed output fields \texttt{answer}, \texttt{claims}, and \texttt{scope}. These inputs identify the fields modified by the post-retrieval ablations; they do not establish a separately serialized pre-generation four-state map or its independent effectiveness.

The retained Reader interface can be summarized as follows:
\begin{table}[ht]
\centering
\caption{Observed compact Reader input fields in the retained GPT LoCoMo calls.}
\label{tab:reader_input_fields}
\begin{tabularx}{\linewidth}{lX}
\toprule
Field & Role in the Reader input \\
\midrule
\texttt{records} & Selected evidence records with source identifiers, speaker, timestamp, original text, and retrieval rank where available. \\
\texttt{dialogue\_blocks} & Reader-facing grouped dialogue context supplied alongside the selected records; its removal is tested in the post-retrieval field ablation. \\
\texttt{reading\_guide} & Reader-facing instructions for interpreting the evidence. The retained Caroline example documents instructions to preserve session/turn order, retain retrieval ranks, and treat missing turn numbers as gaps rather than adjacency. \\
\bottomrule
\end{tabularx}
\end{table}
This field-level description documents the compact interface, not a separate explicit four-state support object. The field ablations test the effect of evidence organization and guidance while retaining the original evidence records.

\section{Prompt Contracts and Examples}
\label{app:prompt_contract}

BiDE-Mem uses two structured LLM-based transformations: memory assertion construction and the QueryFrame implementation of Question Demands. Instead of relying on unconstrained generation, both transformations follow fixed contracts that define their inputs, outputs, and generation restrictions. These contracts keep the resulting representations traceable and compatible with downstream retrieval and evidence construction.

\subsection{Memory Writer V3 Contract}
\label{app:memory_writer_contract}

The Memory Writer transforms raw interaction records into structured memory assertions. Its objective is to preserve searchable information required by the demand-aware memory schema while maintaining provenance to the original interaction.

The input consists of an immutable original record together with deterministic clause spans. The output is a structured assertion object used to construct the hierarchical memory representation described in Appendix~\ref{app:memory_assertion}.

\begin{table}[ht]
\centering
\caption{Memory Writer V3 contract.}
\label{tab:memory_writer_contract}
\begin{tabularx}{\linewidth}{lX}
\toprule
Component & Description \\
\midrule
Input & Original interaction and clause spans \\
Output & Structured memory assertion object \\
Atomic unit & Clause-level proposition \\
Required information & Entity, relation, value, role, polarity, time \\
Traceability & Source clause provenance \\
Restriction & No unsupported cross-record summaries \\
\bottomrule
\end{tabularx}
\end{table}

The Memory Writer follows several constraints:
\begin{itemize}
\item every clause is processed;
\item coordinated propositions are split into independent assertions;
\item negation, preference, causal, result, temporal, and state information are preserved;
\item each assertion remains linked to its source clause;
\item unsupported inference is prohibited.
\end{itemize}

The executable compiler distinguishes asserted, non-propositional, and unresolved L1 units. An unresolved unit does not become a supported assertion merely because its record has been processed. Assertion and clause identifiers provide the links used to construct retrieval views. The compiler also performs schema normalization and records repairs, including normalization of local identifiers and provenance fields. These operations concern representation validity; they do not provide an independent semantic check of every extracted assertion. Source text remains the reference for interpreting generated memory content.

\subsection{QueryFrame V2 Compiler Contract}
\label{app:queryframe_contract}

The QueryFrame compiler implements the Question Demand representation described in Section~\ref{sec:bide_demands} and Appendix~\ref{app:queryframe}. It transforms an input question into the structured fields used to instantiate $D(Q)$ during inference.

The compiler only encodes question requirements. It does not perform memory retrieval, evidence construction, or answer generation.

The compiler follows the following restrictions:
\begin{itemize}
\item it does not answer the input question;
\item it does not access memory records or retrieval results;
\item it does not use reference answers or gold demand annotations;
\item it does not introduce external knowledge;
\item it only represents information requirements implied by the question.
\end{itemize}
The generated QueryFrame is then used as the implementation-level interface between Question Demands and downstream demand-guided retrieval and evidence construction.

\paragraph{Validation and retry behavior.}

The current retrieval-query compiler parses JSON, checks the required key set and query-kind enumeration, and checks list-valued fields. Relation queries, value queries, required evidence acts, graph operations, and lexical expansions must be nonempty; temporal queries are additionally required for a temporal query kind. When a parsed response violates these checks and attempts remain, the compiler requests a corrected JSON object using the question, previous response, and validation errors. The attempt budget is configurable, with a default of three in this compiler. A terminal schema failure is recorded as uncertain; a terminal request or parsing exception is recorded as an error. This default does not establish the retry count used by every archived run.

Resumable execution stores completed answers and errors separately.

The LongMemEval code also exposes a separately configured deterministic question-to-query adapter. It derives retrieval fields from question wording and makes no generative QueryFrame call. Its activation must therefore be distinguished when counting calls or comparing query-generation methods. In either path, structural validation tests the output contract rather than the correctness of the inferred question requirements.

\subsection{Example Transformations}
\label{app:prompt_examples}

This section illustrates the two structured transformations used in BiDE-Mem. The Alice example follows raw memory through assertion and QueryFrame construction. The subsequent Caroline example shows retrieved evidence serialized for the Reader.
\paragraph{Memory assertion construction.}
Given an interaction:
\begin{quote}
Alice stopped enjoying jazz after moving to Boston.
\end{quote}
The Memory Writer produces a structured assertion:
\begin{verbatim}
{
  "subject": "Alice",
  "relation": "preference_change",
  "value": "stopped enjoying jazz",
  "condition": "after moving to Boston",
  "provenance": "source_clause_12"
}
\end{verbatim}
The assertion preserves the state transition, temporal condition, and source provenance required for later retrieval.

\paragraph{QueryFrame construction.}
Given the question:
\begin{quote}
When did Alice stop enjoying jazz?
\end{quote}
The QueryFrame compiler produces:
\begin{verbatim}
{
  "relation_family": "state_change",
  "answer_role": "time",
  "temporal_constraint": "end_time",
  "evidence_shape": "transition"
}
\end{verbatim}
The resulting QueryFrame is the implementation-level realization of the corresponding Question Demands. It guides retrieval toward evidence describing the relevant state transition and its temporal condition rather than only retrieving semantically similar memories.

These two objects are explanatory examples of the conceptual fields. They are not verbatim complete outputs of the current executable contracts described in Appendices~\ref{app:memory_assertion} and~\ref{app:queryframe}.

\paragraph{Recorded evidence serialization.}

A retained LoCoMo example asks, ``When did Caroline go to the LGBTQ support group?'' The following excerpt reproduces one selected record and the packet field names used by the compact Reader. Other selected records are omitted from this excerpt; the example does not represent a one-record evaluation.

{\scriptsize
\begin{verbatim}
{
  "QUESTION": "When did Caroline go to the LGBTQ support group?",
  "ORIGINAL_RAW_PACKET": {
    "records": [{
      "raw_id": "conv-26::D1:3",
      "speaker": "Caroline",
      "source_timestamp": "1:56 pm on 8 May, 2023",
      "original_text": "[quoted below]",
      "image_caption": "",
      "retrieval_rank": 1
    }]
  }
}
\end{verbatim}
}

The original text of the selected record is ``I went to a LGBTQ support group yesterday and it was so powerful.'' The JSON block above abbreviates only this long field to fit the column. The packet also instructs the Reader to follow original session and turn order, retain retrieval ranks, and treat missing turn numbers as gaps rather than adjacency. The timestamp supplies the reference date for ``yesterday'', and the saved answer identifies 7 May 2023. The source identifier and retrieval rank retain provenance independently of display order. This is an observed evidence-to-answer example, not a recorded four-state $Z_Q$ trace.

\section{Supplementary Experimental Results}

\label{app:supplementary}

\subsection{Unified Qwen Evaluation}

\label{app:qwen_protocol}

Table~\ref{tab:app_qwen_overall} reports supplementary results for Qwen3.8-27B-generated answers evaluated by Qwen3.8-27B and GPT-5.4-mini Judges. LoCoMo contains 1,540 questions and LongMemEval-S contains 500. Overall and category scores use the same evaluated answer set within each reported column.
\begin{table}[htbp]
\centering\small
\caption{Supplementary accuracy (\%) using Qwen3.8-27B-generated answers and Qwen3.8-27B / GPT-5.4-mini Judges. Mnemis on LongMemEval-S uses author-provided memory contexts.}
\label{tab:app_qwen_overall}
\resizebox{\linewidth}{!}{%
\begin{tabular}{lrrrr}
\toprule
& \multicolumn{2}{c}{LoCoMo} & \multicolumn{2}{c}{LongMemEval-S} \\
\cmidrule(lr){2-3}\cmidrule(lr){4-5}
Method & Qwen Judge & GPT Judge & Qwen Judge & GPT Judge \\
\midrule
\textbf{BiDE-Mem} & \textbf{86.43} & \textbf{89.94} & \textbf{84.60} & \textbf{88.40} \\
Mnemis & 84.42 & 87.53 & 78.40 & 84.40 \\
EverMemOS & 82.92 & 86.43 & 77.40 & 80.00 \\
Mem0 & 65.52 & 72.73 & 69.40 & 70.40 \\
MIRIX & 70.91 & 75.84 & 50.80 & 57.80 \\
ReadAgent & 71.82 & 75.97 & 46.20 & 51.00 \\
MemoryBank & 46.30 & 49.09 & 39.60 & 40.80 \\
\bottomrule
\end{tabular}%
}
\end{table}
\paragraph{Overall comparison.}

BiDE-Mem achieves the highest overall accuracy under both Judges on both benchmarks. Relative to Mnemis, the differences are 2.01/2.40 percentage points on LoCoMo and 6.20/4.00 on LongMemEval-S (Qwen/GPT Judge). Memory inputs and retrieval procedures remain system-specific; in particular, LongMemEval-S Mnemis uses released author contexts.

\subsection{Category-wise Results and Judge Analysis}

\label{app:category_reporting}

\label{app:judge_analysis}

Tables~\ref{tab:app_locomo_types} and~\ref{tab:app_lme_types} report category-wise accuracy for the same overall result selection. Each category is listed twice: Q denotes Qwen Judge and G denotes GPT Judge, with accuracy shown in percent. Category denominators appear in the row labels; overall accuracy is computed from the summed integer correct counts, not from an unweighted average of rounded category percentages. The reported category counts reconcile exactly with all seven system-level totals.
\begin{table*}[t]
\centering

\scriptsize
\setlength{\tabcolsep}{2.0pt}
\renewcommand{\arraystretch}{1.06}
\caption{LoCoMo category-wise accuracy (\%). Rows identify category size and Judge: Q = Qwen, G = GPT. Values are percentages.}
\label{tab:app_locomo_types}
\begin{tabularx}{\linewidth}{>{\raggedright\arraybackslash}p{0.225\linewidth}*{7}{>{\centering\arraybackslash}X}}
\toprule
Category (N; Judge) & \textbf{BiDE} & Mnemis & EverMemOS & Mem0 & MIRIX & ReadAgent & \shortstack{Memory\\Bank} \\
\midrule
Multi-hop (282; Q) & 77.66 & 71.63 & \textbf{82.98} & 56.38 & 71.99 & 69.15 & 30.85 \\
Multi-hop (282; G) & \textbf{84.04} & 76.60 & 82.27 & 68.79 & 75.53 & 74.82 & 34.04 \\
\addlinespace[2pt]
Temporal (321; Q) & \textbf{82.55} & 80.69 & 81.31 & 65.73 & 66.67 & 53.89 & 41.74 \\
Temporal (321; G) & \textbf{87.85} & 84.74 & 86.60 & 72.59 & 71.03 & 60.75 & 42.68 \\
\addlinespace[2pt]
Open-domain (96; Q) & 59.38 & 55.21 & \textbf{64.58} & 46.88 & 53.13 & 54.17 & 19.79 \\
Open-domain (96; G) & 64.58 & 63.54 & \textbf{69.79} & 56.25 & 64.58 & 57.29 & 22.92 \\
\addlinespace[2pt]
Single-hop (841; Q) & \textbf{93.94} & 93.46 & 85.61 & 70.63 & 74.20 & 81.57 & 56.24 \\
Single-hop (841; G) & \textbf{95.60} & 95.01 & 89.66 & 75.98 & 79.07 & 84.30 & 59.57 \\
\bottomrule
\end{tabularx}
\end{table*}
\begin{table*}[t]
\centering

\scriptsize
\setlength{\tabcolsep}{2.0pt}
\renewcommand{\arraystretch}{1.06}
\caption{LongMemEval-S category-wise accuracy (\%). Rows identify category size and Judge: Q = Qwen, G = GPT. Values are percentages.}
\label{tab:app_lme_types}
\begin{tabularx}{\linewidth}{>{\raggedright\arraybackslash}p{0.225\linewidth}*{7}{>{\centering\arraybackslash}X}}
\toprule
Category (N; Judge) & \textbf{BiDE} & Mnemis & EverMemOS & Mem0 & MIRIX & ReadAgent & \shortstack{Memory\\Bank} \\
\midrule
SS-User (70; Q) & \textbf{97.14} & 92.86 & 84.29 & 92.86 & 71.43 & 85.71 & 82.86 \\
SS-User (70; G) & \textbf{100.00} & 98.57 & 85.71 & 94.29 & 80.00 & 85.71 & 84.29 \\
\addlinespace[2pt]
SS-Assistant (56; Q) & 98.21 & \textbf{100.00} & 75.00 & 73.21 & 46.43 & 82.14 & 91.07 \\
SS-Assistant (56; G) & \textbf{100.00} & \textbf{100.00} & 76.79 & 73.21 & 50.00 & 85.71 & 92.86 \\
\addlinespace[2pt]
SS-Preference (30; Q) & 80.00 & 53.33 & \textbf{83.33} & 73.33 & 30.00 & 13.33 & 16.67 \\
SS-Preference (30; G) & \textbf{83.33} & 73.33 & 66.67 & 73.33 & 43.33 & 36.67 & 26.67 \\
\addlinespace[2pt]
Multi-session (133; Q) & 72.93 & 72.18 & \textbf{75.19} & 63.91 & 50.38 & 27.82 & 21.05 \\
Multi-session (133; G) & 77.44 & \textbf{78.20} & 77.44 & 65.41 & 54.14 & 33.83 & 21.05 \\
\addlinespace[2pt]
Update (78; Q) & \textbf{88.46} & \textbf{88.46} & 85.90 & 85.90 & 66.67 & 58.97 & 46.15 \\
Update (78; G) & 94.87 & 94.87 & \textbf{97.44} & 85.90 & 76.92 & 64.10 & 47.44 \\
\addlinespace[2pt]
Temporal (133; Q) & \textbf{82.71} & 67.67 & 70.68 & 50.38 & 37.59 & 28.57 & 15.04 \\
Temporal (133; G) & \textbf{85.71} & 72.93 & 73.68 & 51.88 & 45.11 & 30.83 & 15.04 \\
\bottomrule
\end{tabularx}
\end{table*}
\paragraph{Independent Judge analysis.}

The two-Judge evaluation scores the same Qwen-generated answers with Qwen3.8-27B and GPT-5.4-mini. GPT-Judge overall accuracies exceed Qwen-Judge accuracies for all seven systems in Table~\ref{tab:app_qwen_overall}, while the overall system ordering is unchanged within each benchmark. These aggregate results describe ranking consistency rather than question-level Judge agreement.

For the confirmed LongMemEval-S result selection, BiDE-Mem scores 423/500 (84.60\%) under Qwen Judge and 442/500 (88.40\%) under GPT Judge. Its Preference counts are 24/30 and 25/30 (80.00/83.33\%); EverMemOS scores 83.33/66.67\% and Mnemis scores 53.33/73.33\%. BiDE-Mem leads Temporal under both Judges, while category leaders vary elsewhere.

Question-level agreement and Cohen's $\kappa$ require paired labels for the same saved answers. For $n$ valid pairs, let $n_{ab}$ count Qwen label $a$ and GPT label $b$, with correctness encoded as 1. Agreement is $p_o=(n_{11}+n_{00})/n$, and $\kappa=(p_o-p_e)/(1-p_e)$ uses the empirical label marginals for $p_e$. These paired statistics cannot be recovered from overall or category-wise percentages alone and are not inferred from the accuracy tables.

\subsection{Fine-grained Ablation Study}

\label{app:ablation_configuration}

\label{app:run_identity}

All component comparisons use the unified original-paper LoCoMo ablation setting: 1,540 answerable questions, GPT-5.4-mini Reader, the same evaluation protocol, and a Top20 evidence budget. These experiments are distinct from the supplementary Qwen evaluation. The submitted Full reference is 1,409/1,540 (91.49\%). Each intervention is defined by its changed components, with its corresponding control stated where applicable.

\paragraph{Typed query representation.}

Standard Query Decomposition uses the \texttt{without\_\allowbreak parsing} query-input configuration under the same retrieval stack and Top20 evidence budget. It achieves 1,349/1,540 (87.60\%) versus 1,409/1,540 (91.49\%) for submitted Full, a 3.89-point difference. The configuration changes the query-input representation, including its \texttt{condition} field; its saved inputs do not separately document generated natural-language subquestions. Accordingly, this is a query-configuration comparison, not an isolated parser-only intervention.
\begin{table}[htbp]
\centering\small
\caption{Demand-related results under the original-paper GPT--Top20 setting (percent).}
\label{tab:app_typed_results}
\begin{tabularx}{\linewidth}{XrX}
\toprule
Configuration & Accuracy & Result count / identifier \\
\midrule
Full & 91.49 & 1,409/1,540; original submission \\
Without explicit demand & 78.70 & 1,212/1,540; original submission \\
Standard Query Decomposition & 87.60 & 1,349/1,540; without\_parsing \\
Without demand-conditioned retrieval & 86.69 & 1,335/1,540; retrieval intervention \\
\bottomrule
\end{tabularx}
\end{table}
\paragraph{Retrieval and reranking.}

The four retrieval configurations follow the same GPT--Top20 evaluation protocol. The single-channel pair uses the same dense candidate pool, changing reranking before Top20 selection; this can change both rank and selected evidence membership. The retained Top20 sets overlap by 9.41 records per question on average. Correct answers increase from 1,164 to 1,301, with 181 incorrect-to-correct and 44 correct-to-incorrect transitions (net +137/1,540; +8.90 points). The multichannel results permit the corresponding setting-level comparisons; only the single-channel reranking contrast has independently retained paired outcomes. The original 68.83\% joint removal uses a different fallback configuration.
\begin{table}[htbp]
\centering\small
\caption{Retrieval and reranking results under the original-paper GPT--Top20 setting (percent). The single-channel row also has retained paired outcomes.}
\label{tab:app_retrieval_results}
\begin{tabularx}{\linewidth}{Xrr}
\toprule
Retrieval & No final reranker & With final reranker \\
\midrule
Single dense channel & 75.58 & 84.48 \\
Multiple channels & 84.81 & 91.49 (submitted Full) \\
\bottomrule
\end{tabularx}
\end{table}
\paragraph{Evidence organization and Reader guidance.}

The post-retrieval interventions remove \texttt{dialogue\_blocks}, \texttt{reading\_guide}, or both from the GPT Reader input while retaining the original evidence records. The matched control for joint removal achieves 1,332/1,540 (86.49\%), compared with 1,302/1,540 (84.55\%) without both fields: 30 fewer correct answers (1.95 percentage points). The submitted Full score (91.49\%) is a separate reference. These field comparisons test evidence organization and Reader guidance, not an independently generated four-state $Z_Q$ verifier.
\begin{table}[htbp]
\centering\small
\caption{Post-retrieval field ablations under the original-paper GPT--Top20 setting (percent). The assembly control is matched to joint field removal; Full provides the overall reference.}
\label{tab:app_assembly_results}
\begin{tabularx}{\linewidth}{Xrr}
\toprule
Configuration & Correct /1,540 & Accuracy \\
\midrule
Full & 1,409 & 91.49 \\
Matched assembly control & 1,332 & 86.49 \\
Without dialogue\_blocks & 1,313 & 85.26 \\
Without reading\_guide & 1,324 & 85.97 \\
Without both fields & 1,302 & 84.55 \\
\bottomrule
\end{tabularx}
\end{table}
\paragraph{Configuration controls.}

All comparisons retain the common dataset, Reader, Judge protocol, and Top20 budget. In the paired single-channel experiment, the query and candidate pool are fixed, while reranking changes which records enter Top20. Reader-input interventions retain the original evidence records; joint field removal uses its matched 86.49\% control. The query-input comparison changes the fields specified above without changing the evaluation protocol.

\section{Computational Cost and Efficiency}

\label{app:cost}

\subsection{Cost Accounting Protocol}

\label{app:cost_protocol}

Online execution begins with the evaluation question and ends with the final answer. The numerical measurements below describe the identified saved execution paths. They describe measured component usage rather than matched end-to-end accuracy--latency measurements. Memory construction and independent judging are separate phases. For each phase, generative calls and input/output tokens are distinguished from embedding, retrieval, reranking, and database operations. Provider-reported token usage and local tokenizer estimates retain separate measurement labels. Missing usage is unavailable rather than zero, and reasoning tokens already included in output usage are not added a second time.

Logical execution cost counts the model operations required by the algorithm, including selection or agent iterations on the successful path. Transport retries, diagnostic reruns, and historical spending are tracked separately. Counting only the final Reader request omits required upstream generative operations for methods that use iterative selection or tool-mediated reasoning.

\subsection{BiDE-Mem Online Cost}

\label{app:bide_online_cost}

For the measured LoCoMo compact path with generative QueryFrame construction, the normal successful path has one QueryFrame call and one Reader call. Retrieval, local reranking, and deterministic evidence organization do not add generative calls. This count excludes memory construction, judgment, and failed attempts. It also excludes the separate explicit ALIGN/VERIFY configuration: when those operations are executed, their additional calls, tokens, and latency must be counted. The formal definition of $Z_Q$ does not establish the call count of a run. A configured deterministic query adapter has a different call structure, as described in Appendix~\ref{app:queryframe_contract}.

Shared saved responses introduce a second accounting perspective. In the recovered LoCoMo records, 1,529 distinct QueryFrame responses are associated with 1,540 questions. A logical per-question view counts the QueryFrame operation for each question; an allocated stored-response view counts each distinct saved response once and distributes it across its associated questions. Both views measure token usage and call allocation, rather than end-to-end latency. Table~\ref{tab:app_bide_tokens} reports both views using the recovered API usage for QueryFrame and accepted Reader responses.

\begin{table}[ht]
\centering\small
\caption{LoCoMo QueryFrame and accepted Reader usage per question over 1,540 questions. These measurements concern the saved Qwen compact-path records; they exclude offline construction and judging. Totals are computed before rounding.}
\label{tab:app_bide_tokens}
\begin{tabularx}{\linewidth}{Xrrr}
\toprule
Accounting unit & Calls/Q & Input tokens/Q & Output tokens/Q \\
\midrule
QueryFrame, logical view & 1 & 447.16 & 290.15 \\
Accepted Reader & 1 & 2,293.48 & 59.12 \\
Logical online total & 2 & 2,740.65 & 349.27 \\
Allocated stored-response total & 1.993 & 2,737.47 & 347.23 \\
\bottomrule
\end{tabularx}
\end{table}

\subsection{Retrieval, Reranking and Baseline Costs}

\label{app:local_cost}

\label{app:system_cost_paths}

Batch wall time divided by the number of questions is an amortized processing cost. It differs from an individual request's elapsed time and from end-to-end online latency. In particular, the retrieved LoCoMo timing covers five measured retrieval stages, not a complete timing of all six fusion families. Their summed amortized time is 0.59 seconds per question. Reranking processes 128 pairs per question on an NVIDIA A100 80GB PCIe, with batch size 8 and a 2,048-token input limit; its batch elapsed time corresponds to 4.35 seconds per question. The accepted Reader requests average 5.24 seconds each in the measured Qwen run. Adding these quantities would not produce a measured end-to-end latency.

For LongMemEval-S, the measured full-text run records amortized times of 7.87, 15.58, 8.57, and 7.21 seconds per question for Dense, Clause, Typed-L2, and final reranking, respectively. The accepted Reader requests average 14,118.70 input tokens and 188.14 output tokens, with mean request time 63.50 seconds. These stages use prepared memory and query inputs, and their measurements do not establish hardware- and load-matched end-to-end latency comparisons.